\documentclass[letterpaper]{article} 
\usepackage{aaai21}  
\usepackage{times}  
\usepackage{helvet} 
\usepackage{courier}  
\usepackage[hyphens]{url}  
\usepackage{graphicx} 
\usepackage{natbib}  
\usepackage{caption} 
\usepackage[switch]{lineno}
\usepackage{amsfonts}
\usepackage{amsmath}
\usepackage{bm}
\usepackage{booktabs} 
\usepackage{color}

\newcommand{\ubold}[1]{\fontseries{b}\selectfont#1}
\newcommand{\ak}[1]{{\color{red}}}

\title{TabTransformer: Tabular Data Modeling \\ Using Contextual Embeddings}
\author{
        Xin Huang,\textsuperscript{\rm 1}
        Ashish Khetan, \textsuperscript{\rm 1}
        Milan Cvitkovic \textsuperscript{\rm 2} 
        Zohar Karnin \textsuperscript{\rm 1} \\
}
\affiliations{
    \textsuperscript{\rm 1} Amazon AWS\\
    \textsuperscript{\rm 2} PostEra \\
    xinxh@amazon.com, 
    khetan@amazon.com,
    mwcvitkovic@gmail.com,
    zkarnin@amazon.com
}

\begin{document}
\maketitle

\begin{abstract}
We propose TabTransformer, a novel deep tabular data modeling architecture for supervised and semi-supervised learning. The TabTransformer is built upon self-attention based Transformers.
The Transformer layers transform the embeddings of categorical features into robust contextual embeddings to achieve higher prediction accuracy.
Through extensive experiments on fifteen publicly available datasets,
we show that 
the TabTransformer outperforms the
state-of-the-art deep learning methods for tabular data
by at least $1.0\%$ on mean AUC, and matches the performance of tree-based ensemble models.
Furthermore, 
we demonstrate that the contextual embeddings learned from TabTransformer are highly robust against both missing and noisy data features, and provide better interpretability.
Lastly, for the semi-supervised setting we develop an unsupervised pre-training procedure to learn data-driven contextual embeddings, resulting in an average $2.1\%$ AUC lift over the state-of-the-art methods.
\end{abstract}

\section{Introduction}
Tabular data is the most common data type in many real-world applications such as recommender systems \citep{cheng2016wide}, online advertising \citep{song_autoint_2019}, and portfolio optimization \citep{ban2018machine}. Many machine learning competitions such as Kaggle and KDD Cup are primarily designed to solve problems in tabular domain. 

The state-of-the-art for modeling tabular data is tree-based ensemble methods such as the gradient boosted decision trees (GBDT) \citep{chen2016xgboost, prokhorenkova2018catboost}. This is in contrast to modeling image and text data where 
all the existing competitive models are based on deep learning \citep{sandler2018mobilenetv2, Devlin2019BERTPO}. 
The tree-based ensemble models can achieve competitive prediction accuracy, are fast to train and easy to interpret. These benefits make them highly favourable among machine learning practitioners. However,
the tree-based models have several limitations in comparison to deep learning models. 
(a) They are not suitable for continual training from streaming data, and do not allow efficient end-to-end learning of image/text encoders in presence of multi-modality along with tabular data. 
(b) In their basic form they are not suitable for state-of-the-art semi-supervised learning methods. This is due to the fact that the basic decision tree learner does not produce reliable probability estimation to its predictions  \citep{Tanha2017SemisupervisedSF}.
(c) The state-of-the-art deep learning methods \citep{Devlin2019BERTPO} to handle missing and noisy data features do not apply to them. Also, robustness of tree-based models has not been studied much in literature. 

A classical and popular model that is trained using gradient descent and hence allows end-to-end learning of image/text encoders is multi-layer perceptron (MLP). 
The MLPs usually learn parametric embeddings to encode categorical data features. But due  to  their  shallow  architecture and context-free embeddings,  they  have  the  following limitations: 
(a) neither the model nor the learned embeddings are interpretable; (b) it is not robust against missing and noisy data (Section \ref{subsec: The Robustness}); 
(c) for semi-supervised learning, they do not achieve competitive performance (Section \ref{subsec: semi-supervised-learning}). 
Most importantly, MLPs do not match the performance of tree-based models such as GBDT on most of the datasets \citep{arik2019tabnet}. To bridge this performance gap between MLP and GBDT, researchers have proposed various deep learning models \citep{song_autoint_2019, cheng2016wide, arik2019tabnet, guo_deepfm_2018}. Although these deep learning models achieve comparable prediction accuracy, 
they do not address all the limitations of GBDT and MLP.  
Furthermore, their comparisons are done in a limited setting of a handful of datasets. In particular, in Section~\ref{subsec: supervised-learning} we show that when compared to standard GBDT on a large collection of datasets, GBDT perform significantly better than these recent models. 

In this paper, we propose TabTransformer to address the limitations of MLPs and existing deep learning models, while bridging the performance gap between MLP and GBDT. We establish performance gain of TabTransformer through extensive experiments on fifteen publicly available datasets. 

The TabTransformer is built upon Transformers \citep{vaswani2017attention} to learn efficient contextual embeddings of categorical features. 
Different from tabular domain, the application of embeddings has been studied extensively in NLP. 
The use of embeddings to encode words in a dense low dimensional space is prevalent in natural language processing. Beginning from Word2Vec \citep{rong2014word2vec} with the context-free word embeddings to BERT \citep{Devlin2019BERTPO} which provides the contextual word-token embeddings, embeddings have been widely studied and applied in practice in NLP. In comparison to context-free embeddings, the contextual embedding based models \cite{mikolov2011extensions, huang2015bidirectional, Devlin2019BERTPO} have achieved tremendous success. In particular, self-attention based Transformers \citep{vaswani2017attention} have become a standard component of NLP models to achieve state-of-the-art performance. The effectiveness and interpretability of contextual embeddings generated by Transformers have been also well studied \citep{coenen2019visualizing, brunner2019validity}.

Motivated by the successful applications of Transformers in NLP,
we adapt them 
in tabular domain. 
In particular, TabTransformer applies a sequence of multi-head attention-based Transformer layers on parametric embeddings to transform them into contextual embeddings, bridging the performance gap between baseline MLP and GBDT models. We investigate the effectiveness and interpretability of the resulting contextual embeddings generated by the Transformers. We find that highly correlated features (including feature pairs in the same column and cross column) result in embedding vectors that are close together in Euclidean distance,
whereas no such pattern exists in context-free embeddings learned in a baseline MLP model. We also study the robustness of the TabTransformer against random missing and noisy data. The contextual embeddings make them highly robust in comparison to MLPs. 




Furthermore, many existing deep learning models for tabular data are designed for supervised learning scenario but few are for semi-supervised leanring (SSL). Unfortunately, the state-of-art SSL models developed in computer vision \citep{voulodimos2018deep, kendall2017uncertainties} and NLP \citep{vaswani2017attention, Devlin2019BERTPO} cannot be easily extended to tabular domain. Motivated by such challenges, we exploit pre-training methodologies from the language models and propose a semi-supervised learning approach for pre-training Transformers of our TabTransformer model using unlabeled data. 

One of the key benefits of our proposed method for semi-supervised learning is the two independent training phases: 
a costly pre-training phase on unlabeled data and a lightweight fine-tuning phase on labeled data. 
This differs from many state-of-the-art semi-supervised methods \citep{chapelle2009semi, oliver2018realistic, stretcu_graph_2019} that require a single training job including both the labeled and unlabeled data. 
The separated training procedure benefits the scenario where the model needs to be pretrained once but fine-tuned multiple times for multiple target variables.
This scenario is in fact quite common in the industrial setting as companies tend to have one large dataset (e.g.\ describing customers/products) and are interested in applying multiple analyses on this data. 
To summarize, we provide the following contributions: 
\begin{enumerate}
\item We propose TabTransformer, an architecture that provides and exploits contextual embeddings of categorical features. We provide extensive empirical evidence showing TabTransformer
is superior to both a baseline MLP and recent deep networks for tabular data while matching the performance of tree-based ensemble models (GBDT).

\item We investigate the resulting contextual embeddings and highlight their interpretability, contrasted to parametric context-free embeddings achieved by existing art.

\item We demonstrate the robustness of TabTransformer against noisy and missing data. 


\item We provide and extensively study a two-phase pre-training then fine-tune procedure for tabular data, beating the state-of-the-art performance of semi-supervised learning methods.
\end{enumerate}

\section{The TabTransformer}\label{sec:TabTransformer}
The TabTransformer architecture comprises a column embedding layer, a stack of $N$ Transformer layers, and a multi-layer perceptron. Each Transformer layer \citep{vaswani2017attention} consists of a multi-head self-attention layer followed by a position-wise feed-forward layer. The architecture of TabTransformer is shown below in Figure \ref{fig:architecture}.
\begin{figure}[h]
\centering
\includegraphics[width=0.9\columnwidth]{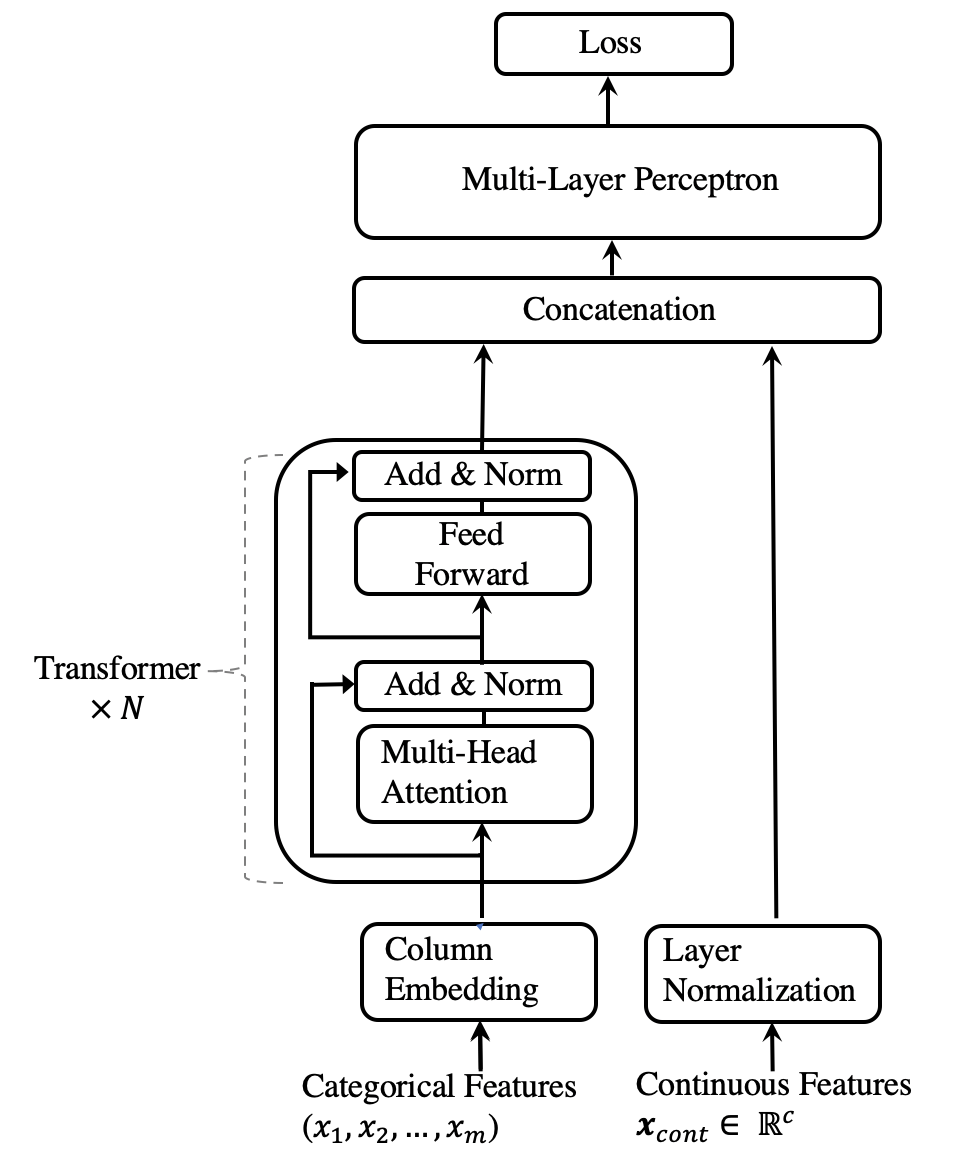} 
\caption{The architecture of TabTransformer.}
\label{fig:architecture}
\end{figure}

Let $(\bm{x}, y)$ denote a feature-target pair,  where $\bm{x} \equiv \{\bm{x}_{\text{cat}}, \bm{x}_{\text{cont}}\}$. The $\bm{x}_{\text{cat}}$ denotes all the categorical features and $\bm{x}_{\text{cont}} \in \mathbb{R}^{c}$ denotes all of the $c$ continuous features. Let $\bm{x}_{\text{cat}} \equiv \{x_1, x_2, \cdots, x_m\}$ with each $x_i$ being a categorical feature, for $i \in \{1,\cdots, m\}$. 

We embed each of the $x_i$ categorical features into a parametric embedding of dimension $d$ using \textit{Column embedding}, which is explained below in detail. Let $\bm{e}_{{\phi}_i}(x_i) \in \mathbb{R}^d$ for $i \in \{1,\cdots,m\}$ be the embedding of the $x_i$ feature, and $\bm{E}_{\phi}(\bm{x}_{\text{cat}}) = \{\bm{e}_{{\phi}_1}(x_1),\cdots,  \bm{e}_{{\phi}_m}(x_m)\}$ be the set of embeddings for all the categorical features. 

Next, these parametric embeddings $\bm{E}_{\phi}(\bm{x}_{\text{cat}})$ are inputted to the first Transformer layer. The output of the first Transformer layer is inputted to the second layer Transformer, and so forth.
Each parametric embedding is transformed into contextual embedding when outputted from the top layer Transformer, through successive aggregation of context from other embeddings. We denote the sequence of Transformer layers as a function $f_{\theta}$. The function $f_{\theta}$ operates on parametric embeddings $\{\bm{e}_{{\phi}_1}(x_1),\cdots, \bm{e}_{{\phi}_m}(x_m)\}$ and returns the corresponding contextual embeddings $\{\bm{h}_1,\cdots, \bm{h}_m\}$ where $\bm{h}_i\in \mathbb{R}^d$ for $i \in \{1,\cdots, m\}$.

The contextual embeddings $\{\bm{h}_1,\cdots, \bm{h}_m\}$ are concatenated along with the continuous features $\bm{x}_{\text{cont}}$ to form a vector of dimension $(d\times m + c)$. This vector is inputted to an MLP, denoted by $g_{\bm{\psi}}$, to predict the target $y$. Let $H$ be the cross-entropy for classification tasks and mean square error for regression tasks. We minimize the following loss function $\mathcal{L}(\bm{x}, y)$ to learn all the TabTransformer parameters in an end-to-end learning by the first-order gradient methods. The TabTransformer parameters include $\bm{\phi}$ for column embedding, $\bm{\theta}$ for Transformer layers, and $\bm{\psi}$ for the top MLP layer.
\begin{align}
\label{equaton: supervised-loss}
\mathcal{L}(\bm{x}, y) \equiv H( g_{\bm{\psi}}(f_{\bm{\bm{\theta}
}}(\bm{E}_{\phi}(\bm{x}_{\text{cat}})), \bm{x}_{\text{cont}}), y)\,.
\end{align}
Below, we explain the Transformer layers and column embedding.

\subsubsection{Transformer.} A Transformer \citep{vaswani2017attention} consists of a multi-head self-attention layer followed by a position-wise feed-forward layer, with element-wise addition and layer-normalization being done after each layer.
A self-attention layer comprises three parametric matrices - Key, Query and Value. Each input embedding is projected on to these matrices, to generate their key, query and value vectors. 
Formally, let $K \in \mathbb{R}^{m \times k}$, $Q \in \mathbb{R}^{m \times k}$ and $V \in \mathbb{R}^{m \times v}$ be the matrices comprising key, query and value vectors of all the embeddings, respectively, and $m$ be the number of embeddings inputted to the Transformer, $k$ and $v$ be the dimensions of the key and value vectors, respectively. Every input embedding attends to all other embeddings through a Attention head, which is computed as follows:
\begin{align}
    \text{Attention}(K, Q, V) = A \cdot V,  
\end{align}
where $A = \text{softmax}({(QK^T)}/{\sqrt{k}})$. For each embedding, the attention matrix $A \in \mathbb{R}^{m \times m}$ calculates how much it attends to other embeddings, thus transforming the embedding into contextual one. The output of the attention head of dimension $v$ is projected back to the embedding of dimension $d$ through a fully connected layer, which in turn is passed through two position-wise feed-forward layers. The first layer expands the embedding to four times its size and the second layer projects it back to its original size. 



\subsubsection{Column embedding.} For each categorical feature (column) $i$, we have an embedding lookup table $\bm{e}_{{\phi}_i}(.)$, for $i \in \{1,2,...,m\}$. For $i$th feature with $d_i$ classes, the embedding table $\bm{e}_{{\phi}_i}(.)$ has $(d_{i}+1)$ embeddings where the additional embedding corresponds to a missing value.
The embedding for the encoded value $x_i = j \in [0, 1, 2, .., d_i]$ is $\bm{e}_{{\phi}_i}(j) = [\bm{c}_{\phi_i}, \bm{w}_{\phi_{ij}}]$, where $\bm{c}_{\phi_i} \in \mathbb{R}^\ell, \bm{w}_{\phi_{ij}} \in \mathbb{R}^{d-\ell}$.
The dimension of $\bm{c}_{\phi_i}$, $\ell$, is a hyper-parameter. The unique identifier $\bm{c}_{\phi_i} \in \mathbb{R}^\ell$ distinguishes the classes in column $i$ from those in the other columns. 

The use of unique identifier is new and is particularly designed for tabular data. 
Rather in language modeling, embeddings are element-wisely added with the positional encoding of the word in the sentence. Since, in tabular data, there is no ordering of the features, we do not use positional encodings. 
An ablation study on different embedding strategies is given in Appendix \ref{app:ablation}. The strategies include both different choices for $\ell,d$ and element-wise adding the unique identifier and feature-value specific embeddings rather than concatenating them.

\subsubsection{Pre-training the Embeddings.}
The contextual embeddings explained above are learned in end-to-end supervised training using labeled examples. For a scenario, when there are a few labeled examples and a large number of unlabeled examples, 
we introduce a pre-training procedure 
to train the Transformer layers using unlabeled data. This is followed by fine-tuning of the pre-trained Transformer layers along with the top MLP layer using the labeled data. For fine-tuning, we use the supervised loss defined in Equation \eqref{equaton: supervised-loss}. 

We explore two different types of pre-training procedures, the masked language modeling (MLM) \citep{Devlin2019BERTPO} and the replaced token detection (RTD) \citep{clark_electra_2020}. Given an input $\bm{x}_{\text{cat}} = \{x_1, x_2, ..., x_m\}$, MLM randomly selects $k \%$ features from index $1$ to $m$ and masks them as missing. 
The Transformer layers along with the column embeddings are trained by minimizing cross-entropy loss of a multi-class classifier that tries to predict the original features of the masked features, from the contextual embedding outputted from the top-layer Transformer.

Instead of masking features, RTD replaces the original feature by a random value of that feature. Here, the loss is minimized for a binary classifier that tries to predict whether or not the feature has been replaced. 
The RTD procedure as proposed in \cite{clark_electra_2020} uses auxiliary generator for sampling a subset of features that a feature should be replaced with. The reason they used an auxiliary encoder network as the generator is that there are tens of thousands of tokens in language data and a uniformly random token is too easy to detect. In contrast, (a) the number of classes within each categorical feature is typically limited; (b) a different binary classifier is defined for each column rather than a shared one, as each column has its own embedding lookup table.
We name the two pre-training methods as TabTransformer-MLM and TabTransformer-RTD. In our experiments, the replacement value $k$ is set to $30$. An ablation study on $k$ is given in Appendix \ref{app:ablation}.


\section{Experiments} \label{sec:experiments}

\subsubsection{Data.} We evaluate TabTransformer and baseline models on $15$ publicly available binary classification datasets from the UCI repository \citep{UCI}, the AutoML Challenge \citep{automlchallenges}, and Kaggle \citep{kaggle_inc_state_2017} for both supervised and semi-supervised learning. Each dataset is divided into five cross-validation splits. The  training/validation/testing proportion of the data for each split are $65/15/20\%$. The number of categorical features across dataset ranges from $2$ to $136$. In the semi-supervised experiments, for each dataset and split, the first $p$  observations in the training data are marked as the labeled data and the remaining training data as the unlabeled set. The value of $p$ is chosen as $50$, $200$, and $500$, corresponding to $3$ different scenarios.
In the supervised experiments, each training dataset is fully labeled. Summary statistics of the all the datasets are provided in Table \ref{tab:dataset_info}, \ref{sup:tab:dataset_urls} in Appendix \ref{sup:experiment_results}.

\subsubsection{Setup.} For the TabTransformer, the hidden (embedding) dimension, the number of layers and the number of attention heads are fixed to $32$, $6$, and $8$ respectively.  The MLP layer sizes are set to $\{4\times l, 2\times l\}$, where $l$ is the size of its input.
For hyperparameter optimization (HPO), each model is given $20$ HPO rounds for each cross-validation split.
For evaluation metrics, we use the Area under the curve (AUC) \citep{bradley1997use}.
Note, the pre-training is only applied in semi-supervised scenario. We do not find much benefit in using it when the entire data is labeled. Its benefit is evident when there is a large number of unlabeled examples and a few labeled examples. Since in this scenario the pre-training provides a representation of the data that could not have been learned based only on the labeled examples.

The experiment section is organized as follows. In Section \ref{subsec: The Effectiveness}, we first demonstrate the effectiveness of the attention-based Transformer by comparing our model with the one without the Transformers (equivalently an MLP model). In Section \ref{subsec: The Robustness}, we illustrate the robustness of TabTransformer against noisy and missing data. Finally, extensive evaluation on various methods are conducted in Section \ref{subsec: supervised-learning} for supervised learning, and in Section \ref{subsec: semi-supervised-learning} for semi-supervised learning.

\subsection{The Effectiveness of the Transformer Layers} \label{subsec: The Effectiveness}
First, a comparison between TabTransformers and the baseline MLP is conducted in a supervised learning scenario. We remove the Transformer layers $f_{\bm{\theta}}$ from the architecture, fix the rest of the components, and compare it with the original TabTransformer. The model without the attention-based Transformer layers is equivalently an MLP. The dimension of embeddings $d$ for categorical features is set as $32$ for both models. The comparison results over $15$ datasets are presented in Table \ref{tab:ab test}. The TabTransformer with the Transformer layers outperforms the baseline MLP on $14$ out of $15$ datasets with an average $1.0\%$ gain in AUC. 

\begin{table}
\caption{Comparison between TabTransfomers and the baseline MLP. The evaluation metric is AUC in percentage.}
\centering
\label{tab:ab test}
\setlength{\tabcolsep}{4pt}
\scalebox{0.87}{
\begin{tabular}{lccc}
\toprule
Dataset  &  Baseline MLP &  TabTransformer & Gain (\%) \\
\midrule
albert & 74.0 & 75.7 & \bf 1.7\\

1995\_income   &    90.5   &          90.6  & \bf 0.1\\ 
dota2games & 63.1 & 63.3 & \bf 0.2 \\
hcdr\_main & 74.3 & 75.1 & \bf 0.8\\
adult          &    72.5  &  73.7    & \bf 1.2  \\
bank\_marketing &                         92.9 &                         93.4  & \bf 0.5 \\
blastchar &        83.9 &           83.5  & -0.4 \\ 
insurance\_co &         69.7 &       74.4  & \bf 4.7 \\
jasmine &    85.1 &   85.3 & \bf 0.2\\
online\_shoppers &                        91.9 &                         92.7 & \bf 0.8\\
 philippine &                        82.1 &                         83.4  & \bf 1.3\\
 qsar\_bio &                         91.0 &                        91.8 & \bf 0.8\\
  seismicbumps &                73.5 &                       75.1 & \bf 1.6\\
  shrutime &                         84.6 &                         85.6 & \bf 1.0\\
  spambase &                         98.4 &                        98.5  & \bf 0.1\\
\bottomrule
\end{tabular}}
\end{table}

\begin{figure*}[t]
\centering
\includegraphics[width=0.95\textwidth]{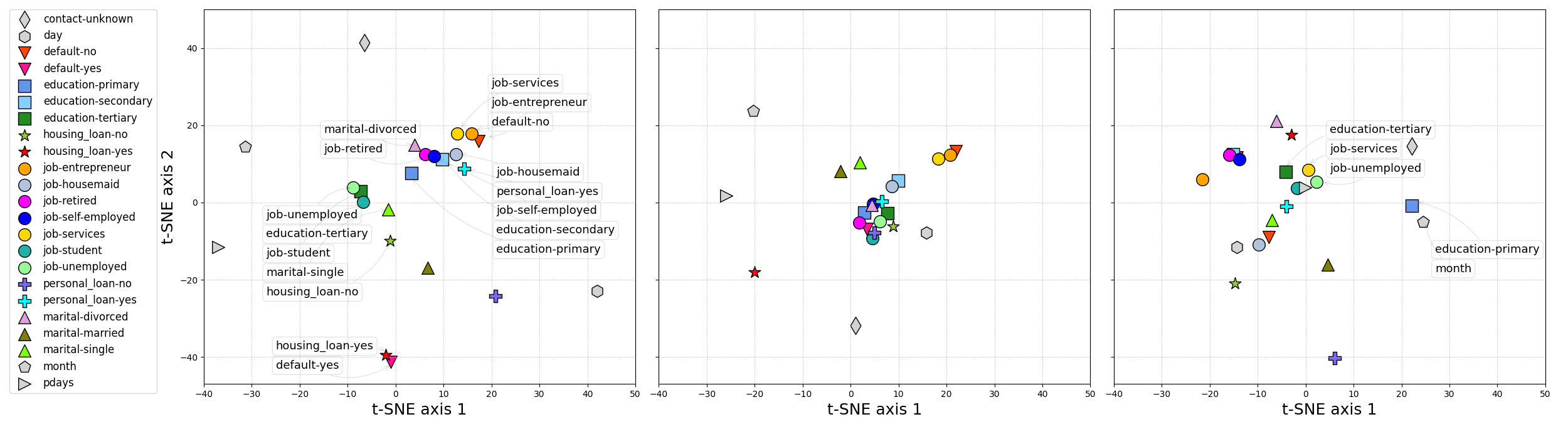} 
\caption{t-SNE plots of learned embeddings for categorical features on dataset \textit{BankMarketing}. \textbf{Left}: TabTransformer-the embeddings generated from the last layer of the attention-based Transformer. \textbf{Center}: TabTransformer-the embeddings before being passed into the attention-based Transformer. \textbf{Right}: The embeddings learned from MLP.}
\label{fig:tsne-embeddings}
\end{figure*}

Next, we take contextual embeddings from different layers of the Transformer and compute a t-SNE plot \citep{maaten2008visualizing} to visualize their similarity in function space. More precisely, for each dataset we 
take its test data, 
pass their categorical features into a trained TabTransformer, and extract all contextual embeddings (across all columns) from a certain layer of the Transformer. The t-SNE algorithm is then used to reduce each embedding to a 2D point in the t-SNE plot. Figure \ref{fig:tsne-embeddings} (left) shows the 2D visualization of embeddings from the last layer of the Transformer for dataset \textit{bank\_marketing}. Each marker in the plot represents an average of 2D points over the test data points for a certain class. We can see that semantically similar classes are close with each other and form clusters in the embedding space. Each cluster is annotated by a set of labels. For example, we find that all of the client-based features (color markers) such as job, education level and martial status stay close in the center and non-client based features (gray markers) such as month (last contact month of the year), day (last contact day of the week) lie outside the central area; in the bottom cluster the embedding of owning a housing loan stays close with that of being default; over the left cluster, embeddings of being a student, martial status as single, not having a housing loan, and education level as tertiary get together; and in the right cluster, education levels are closely associated with the occupation types \citep{eduationjob}.
In Figure \ref{fig:tsne-embeddings}, the center and right plots are t-SNE plots of embeddings before being passed through the Transformer and the context-free embeddings from MLP, respectively. For the embeddings before being passed into the Transformer, it starts to distinguish the non-client based features (gray markers) from the client-based features (color markers). For the embeddings from MLP, we do not observe such pattern and many categorical features which are not semantically similar are grouped together, as indicated by the annotation in the plot.

In addition to prove the effectiveness of Transformer layers, on the test data we take all of the contextual embeddings from each Transformer layer of a trained TabTransformer, use the embeddings from each layer along with the continuous variables as features, and separately fit a linear model with target $y$. Since all of the experimental datasets are for binary classification, the linear model is logistic regression. 
The motivation for this evaluation is defining the success of a simple linear model as a measure of quality for the learned embeddings.

\begin{figure}[h]
\centering
\includegraphics[width=0.9\columnwidth]{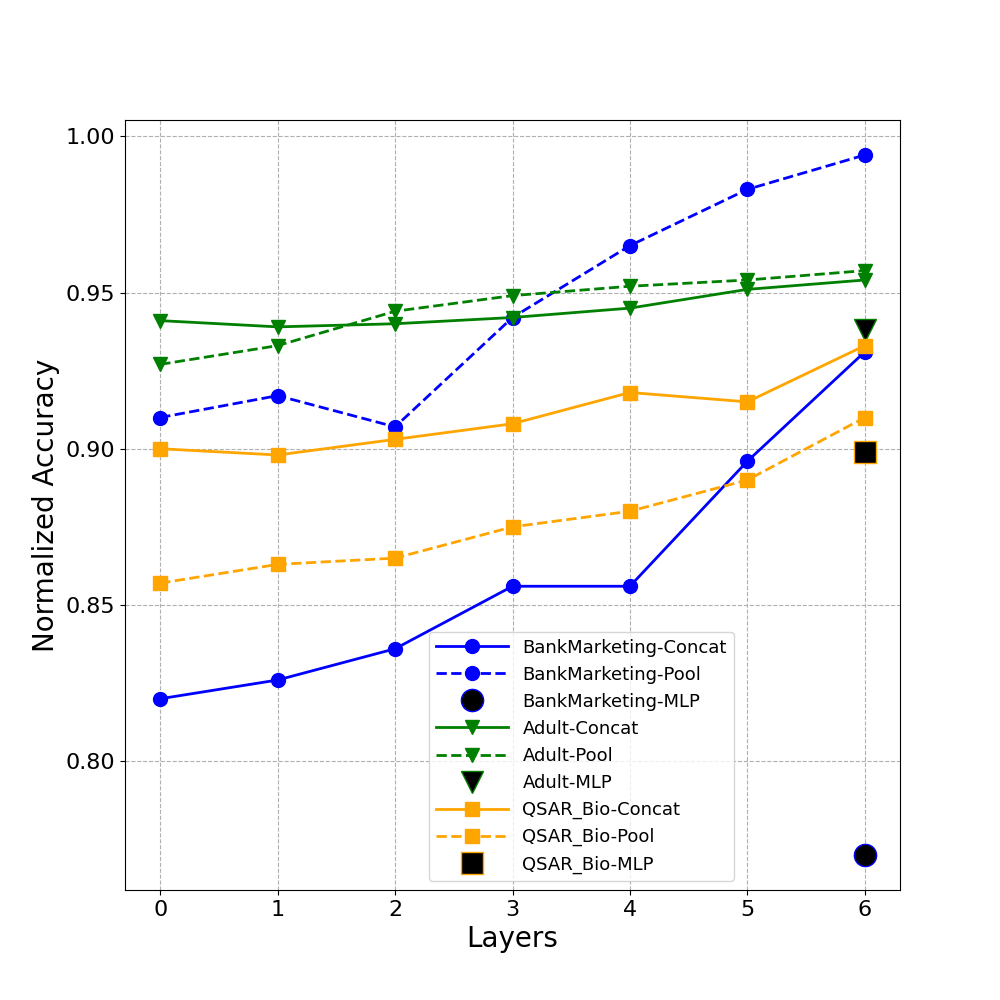} 
\caption{Predictions of liner models using features as the embeddings extracted from different Transformer layers in TabTransformer. Layer $0$ corresponds to the embeddings before being passed into the Transformer layers. For each dataset, each prediction score is normalized by the ``best score" from an end-to-end trained TabTransformer.}
\label{fig:linear-embedding}
\end{figure}

For each dataset and each layer, an average of CV-score in AUC on the test data is computed. The evaluation is conducted on the entire test data with number of data points over 9000. Figure \ref{fig:linear-embedding} presents results for dataset 
\textit{BankMarketing}, \textit{Adult}, and \textit{QSAR\_Bio}. For each line, each prediction score is normalized by the ``best score" from an end-to-end trained TabTransformer for the corresponding dataset. We also explore the average and maximum pooling strategy \citep{howard2018universal} rather than concatenation of embeddings as the features for the linear model. The upward pattern clearly shows that embeddings becomes more effective as the Transformer layer progresses. In contrast, the embeddings from MLP (the single black markers) perform worse with a linear model. Furthermore, the last value in each line close to $1.0$ indicates that a linear model with the last layer of embeddings as features can achieve reliable accuracy, which confirms our assumption. 

\subsection{The Robustness of TabTransformer} \label{subsec: The Robustness}

We further demonstrate the robustness of TabTransformer on the noisy data and data with missing values, against the baseline MLP. We consider these two scenarios only on categorical features to specifically prove the robustness of contextual embeddings from the Transformer layers. 

\subsubsection{Noisy Data.} 
\begin{figure}[h]
\centering
\includegraphics[width=0.9\columnwidth]{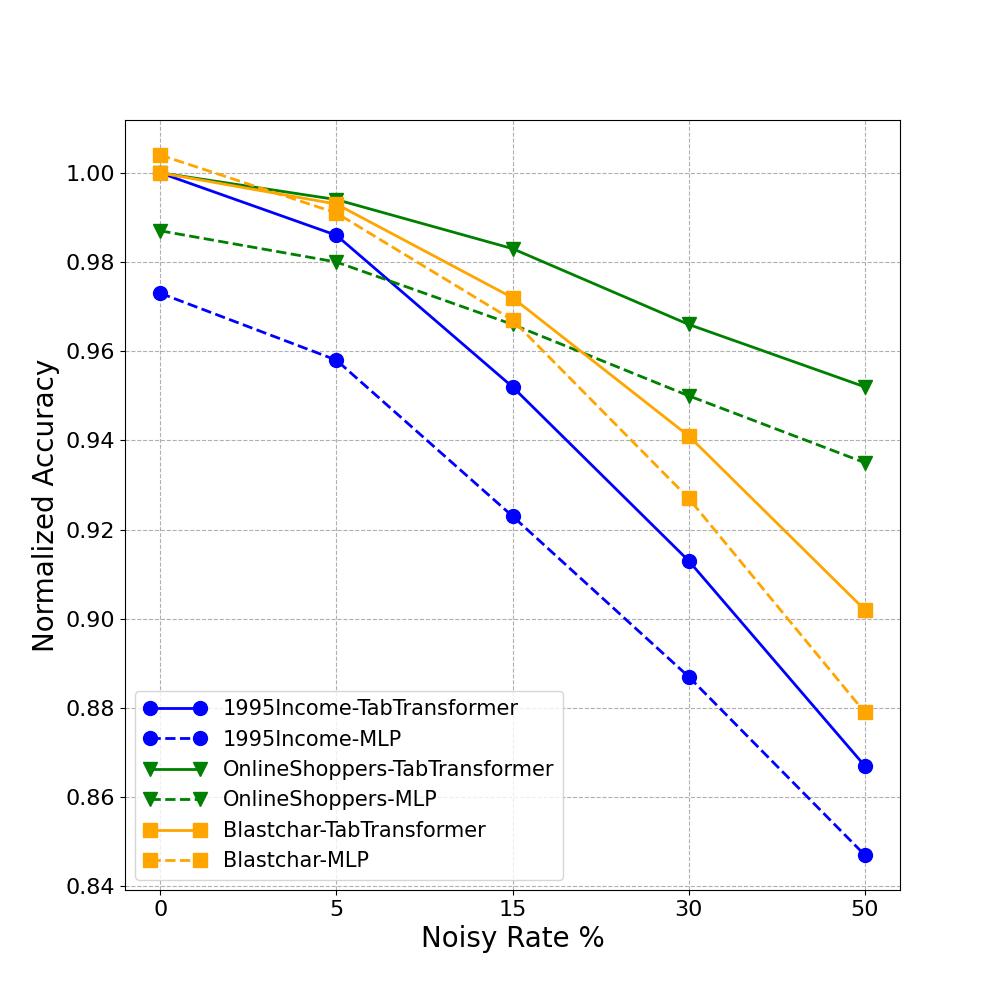} 
\caption{Performance of TabTransformer and MLP with noisy data. For each dataset, each prediction score is normalized by the score of TabTransformer at $0$ noise. 
}
\label{fig:contamination}
\end{figure}

On the test examples, we firstly contaminate the data by replacing a certain number of values by randomly generated ones from the corresponding columns (features). Next, the noisy data are passed into a trained TabTransformer to compute a prediction AUC score. Results on a set of 3 different dataets are presented in Figure \ref{fig:contamination}. As the noisy rate increases, TabTransformer performs better in prediction accuracy and thus is more robust than MLP. In particular notice the \emph{Blastchar} dataset where the performance is near identical with no noise, yet as the noise increases, TabTransformer becomes significantly more performant compared to the baseline.
We conjecture that the robustness comes from the contextual property of the embeddings. Despite a feature being noisy, it draws information from the correct features allowing for a certain amount of correction.

\subsubsection{Data with Missing Values.} 
\begin{figure}[h]
\centering
\includegraphics[width=0.9\columnwidth]{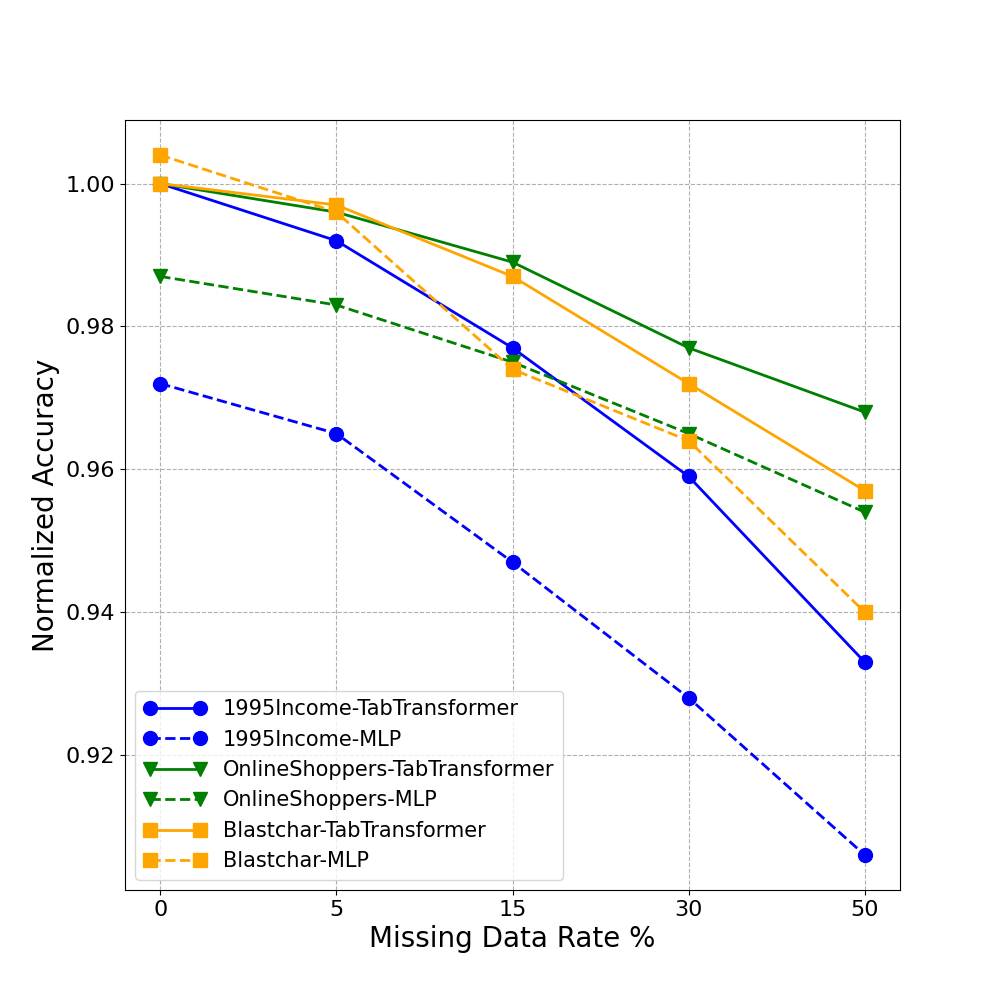} 
\caption{Performance of TabTransformer and MLP under missing data scenario. For each dataset, each prediction score is normalized by the score of TabTransformer trained without missing values. 
}
\label{fig:missing}
\end{figure}

Similarly, on the test data we artificially select a number of values to be missing and send the data with missing values to a trained TabTransformer to compute the prediction score. There are two options to handle the embeddings of missing values: (1) Use the average learned embeddings over all classes in the corresponding column; (2) the embedding for the class of missing value, the additional embedding for each column mentioned in Section ~\ref{sec:TabTransformer}. Since the benchmark datasets do not contain enough missing values to effectively train the embedding in option (2), we use the average embedding in (1) for imputation. Results on the same 3 datasets are presented in Figure \ref{fig:missing}. We can see the same patterns of the noisy data case, i.e.\ that the TabTransformer shows better stability than MLP in handling missing values.

\subsection{Supervised Learning}\label{subsec: supervised-learning}

Here we compare the performance of TabTransformer against following four categories of methods: (a) Logistic regression and GBDT (b) MLP and a sparse MLP following \cite{morcos_one_2019} (c) TabNet model of \citet{arik2019tabnet} (d) and the Variational Information Bottleneck model (VIB) of \citet{alemi_deep_2016}. 

Results are summarized in Table \ref{tab:supervised_result}. TabTransformer, MLP, and GBDT are the top 3 performers. The TabTransformer outperforms the baseline MLP with an average 1.0\% gain and perform comparable with the GBDT. Furthermore, the TabTransformer is significantly better than TabNet and VIB, the recent deep networks for tabular data. For experiment and model details, see Appendix \ref{appendix:model_details}. The models' performances on each individual dataset are presented in Table \ref{tab:supervised-result-auroc1} and \ref{tab:supervised-result-auroc2} in Appendix \ref{sup:experiment_results}. 
\begin{table}
\caption{Model performance in supervised learning. The evaluation metric is mean $\pm$ standard deviation of AUC score over the 15 datasets for each model.
Larger the number, better the result. The top 2 numbers are bold. 
}
\centering
\label{tab:supervised_result}
\setlength{\tabcolsep}{4pt}
\begin{tabular}{lc}
\toprule
Model Name  & Mean AUC (\%) \\
\midrule
TabTransformer & $\mathbf{82.8} \pm 0.4$ \\
MLP & $81.8 \pm 0.4$ \\
GBDT & $\mathbf{82.9}  \pm 0.4$\\
Sparse MLP & $81.4 \pm 0.4$ \\
Logistic Regression & $ 80.4 \pm 0.4$ \\
TabNet & $77.1 \pm 0.5 $ \\
VIB & $80.5 \pm 0.4 $ \\
\bottomrule
\end{tabular}
\end{table}


\subsection{Semi-supervised Learning}\label{subsec: semi-supervised-learning}
Lastly, we evaluate the TabTransformer under the semi-supervised learning scenario where few labeled training examples are available together with a significant number of unlabeled samples. Specifically, we compare our pretrained and then fine-tuned TabTransformer-RTD/MLM against following semi-supervised models: (a) Entropy Regularization (ER) \citep{grandvalet2006entropy} combined with MLP and TabTransformer (b) Pseudo Labeling (PL) \citep{lee2013pseudo} combined with MLP, TabTransformer, and GBDT \citep{JainGBDTPseudolabel} (c)  MLP (DAE): an unsupervised pre-training method designed for deep models on tabular data: the swap noise Denoising AutoEncoder \citep{jahrer_2018}. 

The pre-training models TabTransformer-MLM, TabTransformer-RTD and MLP (DAE)
are firstly pretrained on the entire unlabeled training data and then fine-tuned on labeled data. The semi-supervised learning methods, Pseudo Labeling and Entropy Regularization, are trained on the mix of labeled and unlabeled training data.
To better present results, we split the set of $15$ datasets into two subsets. The first set includes $6$ datasets with more than $30$K data points and the second set includes remaining $9$ datasets.

The results are presented in Table \ref{tab:semi_supervised_result_morethan_30000} and Table \ref{tab:semi_supervised_result_lessthan_30000}. When the number of unlabeled data is large, Table \ref{tab:semi_supervised_result_morethan_30000} shows that our TabTransformer-RTD and TabTransformer-MLM significantly outperform all the other competitors. Particularly, TabTransformer-RTD/MLM improves over all the other competitors by at least $1.2\%$, $2.0\%$ and $2.1\%$ on mean AUC for the scenario of $50$, $200$, and $500$ labeled data points respectively. The Transformer-based semi-supervised learning methods TabTransformer (ER) and TabTransformer (PL) and the tree-based semi-supervised learning method GBDT (PL) perform worse than the average of all the models. When the number of unlabeled data becomes smaller, as shown in Table \ref{tab:semi_supervised_result_lessthan_30000}, TabTransformer-RTD still outperforms most of its competitors but with a marginal improvement.

Furthermore, we observe that when the number of unlabeled data is small as shown in Table \ref{tab:semi_supervised_result_lessthan_30000}, TabTransformer-RTD performs better than TabTransformer-MLM, thanks to its easier pre-training task (a binary classification) than that of MLM (a multi-class classification).
This is consistent with the finding of the ELECTRA paper \citep{clark_electra_2020}. In Table \ref{tab:semi_supervised_result_lessthan_30000}, with only $50$ labeled data points, MLP (ER) and MLP (PL) beat our TabTransformer-RTD/MLM. This can be attributed to the fact that there is room for improvement in our fine-tuning procedure. In particular, our approach allows to obtain informative embeddings but does not allow the weights of the classifier itself to be trained with unlabelled data. Since this issue does not occur for ER and PL, they obtain an advantage in extremely small labelled set. We point out however that this only means that the methods are complementary and mention that a possible follow up could combine the best of all approaches. 


Both evaluation results, Table \ref{tab:semi_supervised_result_morethan_30000} and Table \ref{tab:semi_supervised_result_lessthan_30000}, show that our 
TabTransformer-RTD and Transformers-MLM models are promising in extracting useful information from unlabeled data to help supervised training, and are particularly useful when the size of unlabeled data is large. For model performance on each individual dataset see Table \ref{sup:tab:semisup-result-auroc-50-1}, \ref{sup:tab:semisup-result-auroc-50-2},
\ref{sup:tab:semisup-result-auroc-200-1}, \ref{sup:tab:semisup-result-auroc-200-2},
\ref{tab:semisup-result-auroc-500-1}, \ref{tab:semisup-result-auroc-500-2}
in Appendix \ref{sup:experiment_results}.

\begin{table}[t]
\caption{Semi-supervised learning results for $8$ datasets each with {more than ${30}$K}  data points, for different number of labeled data points. Evaluation metrics are mean AUC in percentage. Larger the number, better the result.}
\centering
\label{tab:semi_supervised_result_morethan_30000}
\setlength{\tabcolsep}{6pt}
\scalebox{0.85}{
\begin{tabular}{lccc}
\toprule
\# Labeled data & $50$ & $200$ & $500$ \\
\midrule
TabTransformer-RTD &   $66.6 \pm 0.6$    & $70.9 \pm 0.6$    &    $\mathbf{73.1} \pm  0.6$  \\
TabTransformer-MLM &    $\mathbf{66.8} \pm 0.6$    &   $\mathbf{71.0} \pm 0.6$  &  $72.9 \pm 0.6$    \\
MLP (ER) & $65.6 \pm 0.6$   & $69.0 \pm 0.6$  & $71.0 \pm 0.6$  \\
MLP (PL) & $65.4 \pm 0.6$  & $68.8 \pm 0.6$ & $71.0 \pm 0.6$   \\
TabTransformer (ER) & $62.7 \pm 0.6$  & $67.1 \pm 0.6$  & $69.3 \pm 0.6$ \\
TabTransformer (PL) & $63.6 \pm 0.6$   & $67.3 \pm 0.7$  &  $69.3 \pm 0.6$ \\
MLP (DAE) &     $65.2 \pm 0.5$      &  $68.5 \pm 0.6$  &   $71.0 \pm 0.6$  \\
GBDT (PL)& $56.5 \pm 0.5$  & $63.1 \pm 0.6$  &  $66.5 \pm 0.7$ \\
\bottomrule
\end{tabular}}
\end{table}

\begin{table}[t]
\caption{Semi-supervised learning results for $12$ datasets each with {less than ${30}$K}  data points, for different number of labeled data points. Evaluation metrics are mean AUC in percentage. Larger the number, better the result.} 
\centering
\label{tab:semi_supervised_result_lessthan_30000}
\setlength{\tabcolsep}{6pt}
\scalebox{0.87}{
\begin{tabular}{lccc}
\toprule
\# Labeled data & $50$ & $200$ & $500$ \\
\midrule
TabTransformer-RTD &   $78.6 \pm 0.6$    & $\mathbf{81.6} \pm 0.5$    &  $ \mathbf{83.4} \pm 0.5$    \\
TabTransformer-MLM &    $78.5 \pm 0.6$   &   $81.0 \pm 0.6$  &    $82.4 \pm 0.5$ \\
MLP (ER) & $\mathbf{79.4} \pm 0.6$   & $81.1 \pm 0.6$   & $82.3 \pm 0.6$ \\
MLP (PL) & $79.1 \pm 0.6$   & $81.1 \pm 0.6$ & $82.0 \pm 0.6$   \\
TabTransformer (ER) & $77.9 \pm 0.6$  & $81.2 \pm 0.6$ &  $82.1 \pm 0.6$  \\
TabTransformer (PL) & $77.8 \pm 0.6$  & $81.0 \pm 0.6$ &  $ 82.1 \pm 0.6$  \\
MLP (DAE) &     $78.5 \pm 0.7$       &  $80.7 \pm 0.6$   &   $82.2 \pm 0.6$  \\
GBDT (PL)& $73.4 \pm 0.7$  & $78.8 \pm 0.6$   &  $81.3 \pm 0.6$ \\
\bottomrule
\end{tabular}}

\end{table}

\section{Related Work}
\textbf{Supervised learning.} Standard MLPs have been applied to tabular data for many years \citep{taxipaper}.
For deep models designed specifically for tabular data, 
there are deep versions of factorization machines \citep{guo_deepfm_2018,xiao_attentional_2017}, Transformers-based methods  \citep{song_autoint_2019,li_interpretable_2020,sun_deepenfm_2019}, and deep versions of decision-tree-based algorithms \citep{ke2019tabnn, yang2018deep}. In particular, \citep{song_autoint_2019} applies one layer of multi-head attention on embeddings to learn higher order features. The higher order features are concatenated and inputted to a fully connected layer to make the final prediction. \citep{li_interpretable_2020} use self-attention layers and track the attention scores to obtain feature importance scores. \citep{sun_deepenfm_2019} combine the Factorization Machine model with transformer mechanism. All 3 papers are focused on recommendation systems making it hard to have a clear comparison with this paper.
%
Other models have been designed around the purported properties of tabular data such as low-order and sparse feature interactions. These include Deep \& Cross Networks \citep{Wang2017DeepC},
Wide \& Deep Networks \citep{cheng2016wide}, TabNets \citep{arik2019tabnet}, and AdaNet \citep{cortes2016adanet}. 

\textbf{Semi-supervised learning.} \cite{izmailov_semi-supervised_2019} give a semi-supervised method based on density estimation and evaluate their approach on tabular data. 
{\em Pseudo labeling} \citep{lee2013pseudo} is a simple, efficient and popular baseline method. 
The Pseudo labeling uses the current
network to infer pseudo-labels of unlabeled examples, by choosing the most confident class. These pseudo-labels are treated like human-provided labels in the cross entropy loss. 
{\em Label propagation}
\citep{zhur2002learning},
\citep{iscen2019label}
is a similar approach where a node’s labels propagate to all nodes according to their proximity,
and are used by the training model as if they were the true labels. Another standard method in semi-supervised learning is {\em entropy regularization}
\citep{grandvalet2005semi, sajjadi2016regularization}. It adds average per-sample entropy for the unlabeled examples to the original loss function for the labeled
examples. Another classical approach of semi-supervised learning is co-training \citep{nigam2000analyzing}. However, the recent approaches - entropy regularization and pseudo labeling - are typically better and more popular. A succinct review of semi-supervised learning methods in general can be found in \citep{oliver_realistic_2019, chappelle2010semi}.

\section{Conclusion}
We proposed TabTransformer, a novel deep tabular data modeling architecture for supervised and semi-supervised learning. We provide extensive empirical evidence showing TabTransformer significantly outperforms MLP and recent deep networks for tabular data while matching the performance of tree-based ensemble models (GBDT). 
We provide and extensively study a two-phase pre-training then fine-tune procedure for tabular data, beating the state-of-the-art performance of semi-supervised learning methods.
TabTransformer shows promising results for robustness against noisy and missing data, and interpretability of the contextual embeddings. For future work, it would be interesting to investigate them in detail. 

%

%
\newpage
\clearpage
\begin{quote}
\begin{small}
\bibliography{reference}
\end{small}
\end{quote}

\newpage
\clearpage


\newpage
\clearpage
\appendix

\section{Appendix: Ablation Studies}
\label{app:ablation}

We perform a number of ablation studies on various architectural choices and pre-training approaches for our TabTransformer. The first ablation study is on the choice of column embedding. The second and third ablation studies focus on the pre-training approach. Specifically, they are on the replacement value $k$ and dynamic versus static replacement strategy. For the pre-training approach, we use TabTransformer-RTD as our model. That is, the loss in the pre-training is RTD loss. For TabTransformer, the hidden (embedding) dimension, the number of layers and the number of attention heads in the Transformer are set to $32$, $6$, and $8$ respectively.  The MLP layer sizes are set to $\{4\times l, 2\times l\}$, where $l$ is the size of its input. To better present the result, we introduce an additional evaluation metric, the \textit{relative AUC}. More precisely, for each dataset and cross-validation split, the \textit{relative AUC} for a model is the relative change of its AUC against the mean AUC over all competing models.

\subsubsection{Column Embedding.} The first study is on the choice of column embedding -- shared parameters $\bm{c}_{\phi_i}$ across the embeddings of multiple classes in column $i$ for $i \in \{1,2, ..., m\}$. In particular, we study the optimal dimension of $\bm{c}_{\phi_i}$, $\ell$. An alternative choice is to element-wisely add the unique identifier $\bm{c}_{\phi_i}$ and feature-value specific embeddings $\bm{w}_{\phi_{ij}}$ rather than concatenating them. In that case, both the dimension of $\bm{c}_{\phi_i}$ and $\bm{w}_{\phi_{ij}}$ are equal to the dimension of embedding $d$.  The goal of having column embedding is to enable the model to distinguish the classes in one column from those in the other columns. A baseline approach is to not have any shared embedding. Results are presented in Table \ref{tab:RTD-layer-number-experiment-mean-normalized-score} where ``Col Embed-Concat-1/$X$" indicates that the dimension $\ell$ is set as $d/X$. The relative AUC score is calculated over all the models that appear in the rows and columns in the table, which explains why negative scores appear in some of the entries. Results show that not having the shared column embedding performs worst and our concatenation column embedding gives an average better performance.


\begin{table*}
\caption{Performance of TabTransformer with no column embedding, concatenation column embedding, and addition column embedding. The evaluation metric is mean $\pm$ standard deviation of relative AUCs (in percentage) over all 15 datasets. Larger value means better performance. The best model is bold for each row.}
\centering
\label{tab:RTD-layer-number-experiment-mean-normalized-score}
\scalebox{0.85}{
\begin{tabular}{crrrr}
\toprule
\# of Transformers Layers&  No Col Embed &  Col Embed-Concat-1/4 &  Col Embed-Concat-1/8 &  Col Embed-Add \\
\midrule
1           &           -0.59 $\pm$ 0.33 &                   -2.01 $\pm$ 1.33 &               {\bf-0.27} $\pm$ {\bf0.21}&            -1.11 $\pm$ 0.77 \\
2           &          -0.59 $\pm$ 0.22 &                   -0.37 $\pm$ 0.20 &                   -0.14 $\pm$ 0.19 &           {\bf0.34} $\pm$ {\bf0.27} \\
3           &           -0.37 $\pm$ 0.19 &                    0.04 $\pm$ 0.18 &                   -0.02 $\pm$ 0.21 &           {\bf0.21} $\pm$ {\bf0.23} \\
6           &            0.54 $\pm$ 0.22 &                    0.53 $\pm$ 0.24 &               {\bf0.70} $\pm$ {\bf0.17} &             0.25 $\pm$ 0.23 \\
12          &            0.66 $\pm$ 0.21 &               {\bf1.05} $\pm$ {\bf0.31} &                    0.73 $\pm$ 0.58 &             0.42 $\pm$ 0.39 \\
\bottomrule
\end{tabular}}
\end{table*}

\subsubsection{The replacement value $k$.} 
\begin{figure*}[t]
\centering
\includegraphics[scale=0.7]{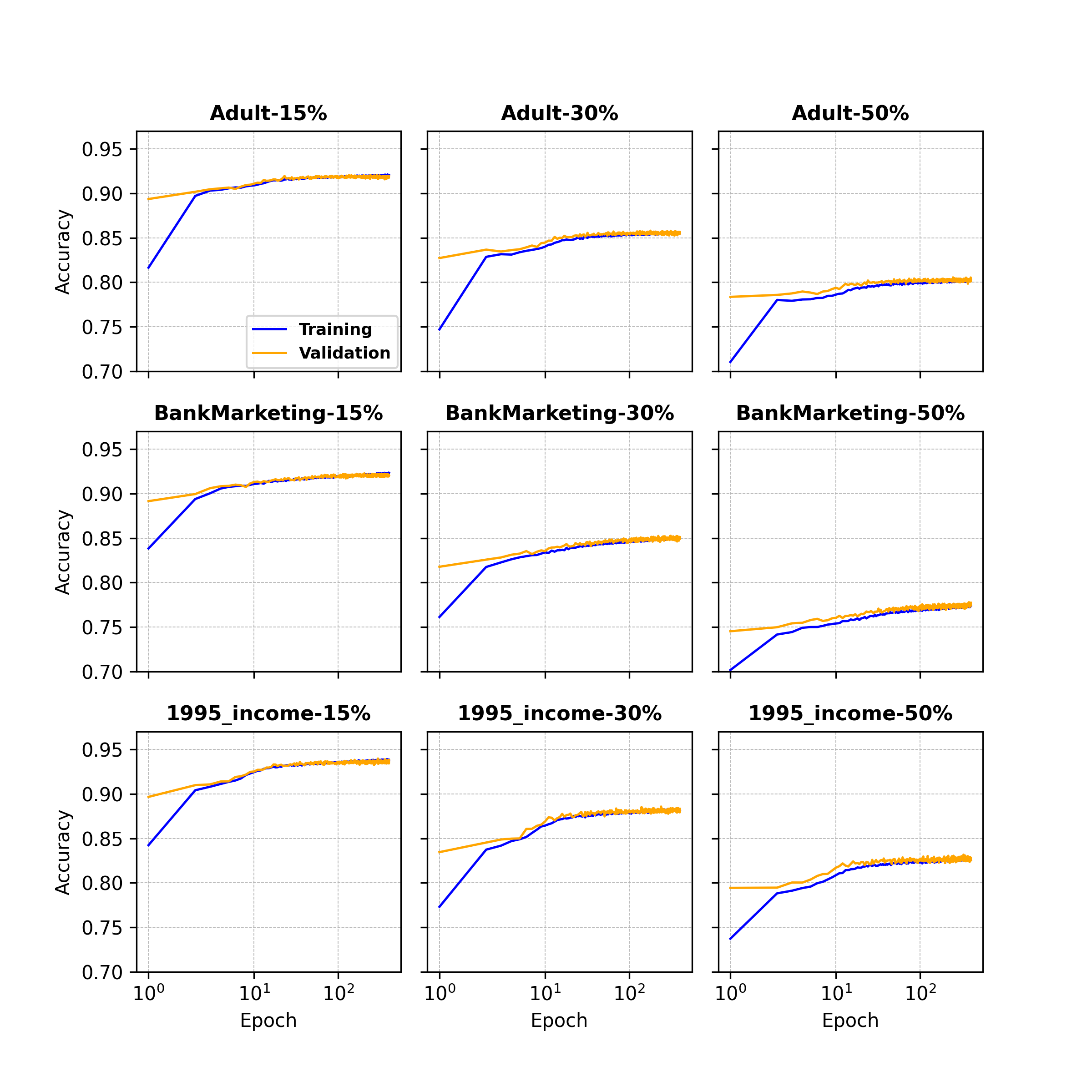}
\caption{The pre-training curves of training and validation accuracy for the three different replacement value $k$ for dataset \textit{Adult}, \textit{BankMarketing}, and \textit{1995\_income}.
}
\label{fig:dynamic_static_plot_log1}
\end{figure*}

The second ablation study is on the replacement value $k$ in pre-training approach. We run experiments for three different choices of $k$ -- $\{15, 30, 50\}$ on three different datasets, namely -- \textit{Adult}, \textit{BankMarketing}, and \textit{1995\_income}. The TabTransformer is firstly pre-trained with a value of $k$ on unlabeled data and then fine-tuned on labeled data. The number of labeled data is set as 256. The final fine-tuning accuracy is not much sensitive to the value of $k$, as shown in Table \ref{tab:retraining-acc-replace-token-rate1}. The pre-training curves of training and validation accuracy for the three different replacement value $k$ is shown in Figure \ref{fig:dynamic_static_plot_log1}. Note, that a constant prediction model would achieve $85\%$ accuracy for the $15\%$ replacement value.

\begin{table*}
\caption{Fine-tuning performance of TabTransformer-RTD for different pre-training replacement value $k$. The number of labeled data points is $256$. The evaluation metrics are mean $\pm$ standard deviation of (1) AUC score over $5$ cross-validation splits for each dataset (in percentage); (2) relative AUCs over the $3$ datasets (in percentage). Larger value means better performance. The best model is bold for each column.}
\centering
\label{tab:retraining-acc-replace-token-rate1}
\scalebox{0.95}{
\begin{tabular}{crrrr}
\toprule
Replacement value $k\%$ &  Adult & BankMarketing & 1995\_income & relative AUC (\%) \\
\midrule
15 &   58.1 $\pm$ 3.52 &        85.9 $\pm$ 1.62 &       \bf86.8 $\pm$ 1.35 & 0.02 $\pm$ 0.10\\
30 &  \bf58.1 $\pm$ 3.15 &         \bf86.1 $\pm$ 1.58  &       86.7 $\pm$ 1.41 & \ubold 0.08 $\pm$ 0.10\\
50 &  57.9 $\pm$ 3.21 &         85.7 $\pm$ 1.93&       86.7 $\pm$ 1.38 & -0.10 $\pm$ 0.11 \\
\bottomrule
\end{tabular}
}
\end{table*}

\subsubsection{Dynamic versus Static Replacement.} 
\begin{figure*}[t]
\centering
\includegraphics[scale=0.7]{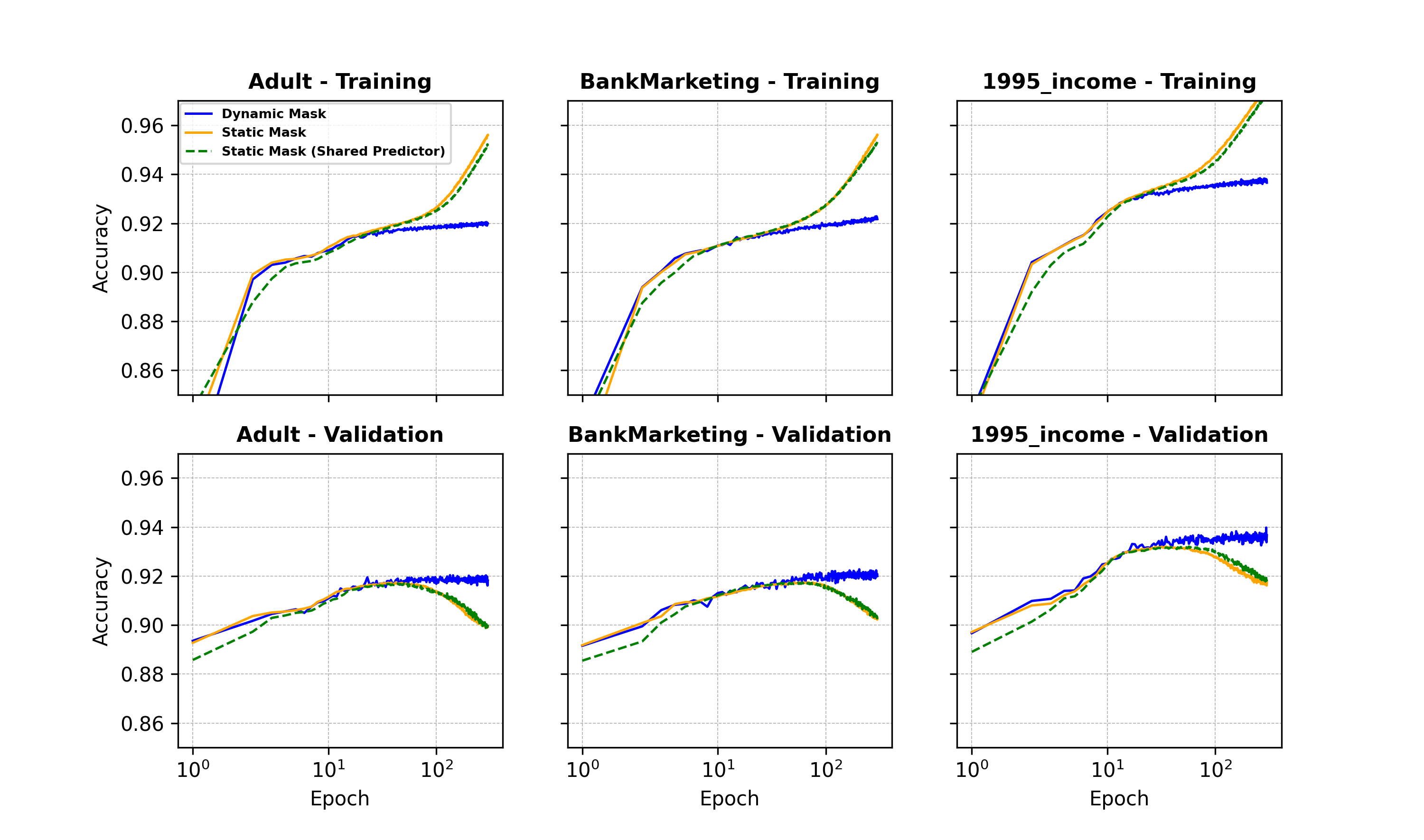}
\caption{The pre-training curves of training and validation accuracy for dynamic mask, static mask, and static mask with a shared predictor (classifier) for dataset \textit{Adult}, \textit{BankMarketing}, and \textit{1995\_income}.
}
\label{fig:dynamic_static_plot_log2}
\end{figure*}

The third ablation study is on dynamic vs static replacement in the pre-training approach. In dynamic replacement, we randomly replace feature values during pre-training over the epochs. That is the replacement is different in each epoch. Whereas in static replacement, the random replacement is chosen once, and then the same replacement is used in all the epochs. We combine this study with another ablation on shared RTD binary classifier (predictor) vs. different classifiers for different columns. Results in Table \ref{tab:RTD-dynamic-static-mask-finetune-result} show that our choice of dynamic replacement and un-shared RTD classifiers perform better than static replacement and shared RTD classifiers. Figure \ref{fig:dynamic_static_plot_log2} shows the pre-training curves of training and validation accuracy for the three choices -- dynamic replacement, static replacement, and static replacement with a shared RTD classifier.

\begin{table*}
\caption{Fine-tuning performance of TabTransformer-RTD for dynamic replacement, static replacement, and static replacement with a shared classifier. The number of labeled data points is $256$. The evaluation metrics are mean $\pm$ standard deviation of (1) AUC score over $5$ cross-validation splits for each dataset (in percentage) ; (2) relative AUCs over the $3$ datasets (in percentage). Larger value means better performance.  The best model is bold for each column. 
}
\centering
\label{tab:RTD-dynamic-static-mask-finetune-result}
\scalebox{0.8}{
\begin{tabular}{lrrrr}
\toprule
{} &    Adult &  BankMarketing & 1995\_income  & relative AUC (\%)\\
\midrule
Dynamic Replacement (Un-shared RTD classifiers)  &  \bf 58.1 $\pm$ 3.52 &        \bf85.9 $\pm$ 1.62 &       \bf86.8 $\pm$ 1.35 & \ubold 0.81 $\pm$ 0.19\\
Static Replacement  (Un-shared RTD classifiers)                         &  57.9 $\pm$ 2.93 &           83.9 $\pm$ 1.18 &        85.9 $\pm$ 1.60 & -0.33 $\pm$ 0.15\\
Static Replacement (Shared RTD Classifiers)       &  57.5 $\pm$ 2.74  &           84.2 $\pm$ 1.46 &        86.0 $\pm$ 1.69 & -0.49 $\pm$ 0.11\\
\bottomrule
\end{tabular}
}
\end{table*}

\section{Appendix: Experiment and Model Details}\label{appendix:model_details}





In this section, we discuss the experiments and model details. First, we go through the experiments details and hyper parameters search space for HPO in Section \ref{sup:model_details_HPO}. Next, we discuss the feature engineering in Section \ref{sup:model_details_feature_engineer}.


\subsection{Experiments Details and Hyper Parameters}\label{sup:model_details_HPO}

\subsubsection{Setup.} All experiments were run on an Ubuntu Linux machine with 8 CPUs and 60GB memory, with all models using a single NVIDIA V100 Tensor Core GPU. For the competing models mentioned in the experiment, we re-implemented all of them for consistency of pre-processing. In cases where there exist published results for a model, our tested results are close to the published records. The GBDT model is implemented using the \texttt{LightGBM} library \citep{ke2017lightgbm}. All the other models are implemented using the \texttt{PyTorch} library \citep{pytorchpaper}. To reproduce our experiment results, the models' implementations and the exact values for all hyper-parameters can be found in another supplemental material, Code and Data Appendix.

For each dataset, all of the cross-validation splits, labeled, and unlabeled training data are obtained with a fixed random seed such that every model tested receives exactly the same training and testing conditions.

As all the datasets are for binary classification, the cross entropy loss was used for both supervised and semi-supervised training (for pre-training, the problem is binary classification in RTD and multi-class classification in MLM). For all deep models, the AdamW optimizer \citep{adamw} was used to update the model parameters, and a constant learning rate was applied throughout each training job. All models used early stopping based on the performance on the validation set and the early stopping patience (the number of epochs) is set as 15.

\subsubsection{Hyper-parameters Search Space.} The hyper-parameters tuned for the GBDT model were the number of leaves in the trees with a search space $\{ x \in \mathbb{Z} \vert 5 \leq x \leq 50\}$, the minimum number of datapoints required to split a leaf in the trees with a search space $\{ x \in \mathbb{Z} \vert 1 \leq x \leq 100\}$, the boosting learning rate with a search space $\{x = 5\cdot 10^u, u \in \mathbb{U} \vert -3 \leq x \leq -1\}$, and the number of trees used for boosting with a search space $\{ x \in \mathbb{Z} \vert 10 \leq x \leq 1000\}$.

For all of the deep models, the common hyper-parameters include the weight decay factor with a search space $\{ x = 10^u, u \in \mathbb{U} \vert  -6 \leq u \leq -1  \}$, the learning rate with a search space $\{  x = 10^u, u \in \mathbb{U} \vert -6 \leq u \leq -3  \}$, the dropout probability with a search space $\{0, 0.1, 0.2,...0.5\}$, and whether to one-hot encode categorical variables or train learnable embeddings.

For MLPs, they all used SELU activations \citep{klambauer2017self} followed by batch normalization in each layer, and set the number of hidden layers as 2. The model-specific hyper-parameters tuned were the first hidden layer with a search space $\{x = m*l, m \in \mathbb{Z} \vert 1 \leq m \leq 8\}$ where $l$ is the input size, and the second hidden layer with a search space $\{x = m*l, m \in \mathbb{Z} \vert 1 \leq m \leq 3\}$.


For TabTransformer, the hidden (embedding) dimension, the number of layers and the number of attention heads in the Transformer were fixed to $32$, $6$, and $8$ respectively during the experiments. The MLP layer sizes were fixed to $\{4\times l, 2\times l\}$, where $l$ was the size of its input. However, these parameters were optimally selected based on 50 rounds of HPO run on 5 datasets.  The search spaces were the number of attention heads $\{ 2,4,8 \}$, the hidden dimension $\{ 32, 64, 128, 256 \}$, and the number of layers $\{ 1, 2, 3, 6, 12 \}$. The search spaces of the first and second hidden layer in MLP are exactly the same as those in MLP model setting. The dimension of $\bm{c}_{\phi_i}$, $\ell$ was chosen as $d/8$ based on the ablation study in Appendix \ref{app:ablation}.


For Sparse MLP (Prune), its implementation was the same as the MLP except that at every $k$ epochs during training the fraction $p$ of weights with the smallest magnitude were permanently set to zero.
The model-specific hyper-parameters tuned were the fraction $p$ with a search space $\{ x = 5\cdot10^u, u \in \mathbb{U} \vert -2 \leq u \leq -1  \}$. The number of layers and layer sizes are exactly the same as the setting in MLP. The parameter $k$ is set as 10.

For TabNet model, we implemented exactly as described in \citet{arik2019tabnet}, though we also added the option to use a softmax attention instead of a sparsemax attention, and did not include the sparsification term in the loss function.
The model-specific hyper-parameters tuned were the number of layers with a search space $\{ x \in \mathbb{Z} \vert 3 \leq x \leq 10\}$ , the hidden dimension $\{ x \in \mathbb{Z} \vert 8 \leq x \leq 128\}$, and the sparse coefficient with a search space $\{ x = 10^u, u \in \mathbb{U} \vert  -6 \leq u \leq -2  \}$.

For VIB model, we implemented it as described in \citet{alemi_uncertainty_2018}.  We used a diagonal covariance, with 10 samples from the variational distribution during training and 20 during testing.
The model-specific hyper-parameters tuned were the number of hidden layers and layer sizes, with exactly the same search spaces as MLP, and the number of mixture components in the mixture of gaussians used in the marginal distribution with a search space $\{ x \in \mathbb{Z} \vert 3 \leq x \leq 10\}$.



For MLP (DAE), its pre-training used swap noise as described in \citet{jahrer_2018}. The model-specific hyper-parameters were exactly the same as MLP.

For Pseudo Labeling \citep{lee2013pseudo}, since this method was combined with deep models such as MLP, TabTransformer and GBDT, the model-specific hyper-parameters were exactly the same as the corresponding deep models mentioned above. The unsupervised coefficient $\alpha$ is chosen as $\alpha_{f} = 3, T_1 = 30, T_2 = 70$. 

For Entropy Regularization \citep{grandvalet2006entropy},  it is the same as Pseudo Labeling. The additional model-specific hyper-parameter was the positive Lagrange multiplier $\lambda$ with a search space $\{0.1, 0.2, ..., 0.9\}$.

\subsection{Feature Engineering}\label{sup:model_details_feature_engineer}

For categorical variables, the processing options include whether to one-hot encode versus learn a parametric embedding, what embedding dimension to use, and how to apply dropout regularization (whether to drop vector elements or whole embeddings).
In our experiments we found that learned embeddings nearly always improved performance as long as the cardinality of the categorical variable is significantly less than the number of data points, otherwise the feature is merely a means for the model to overfit.

For scalar variables, the processing options include how to re-scale the variable (via quantiles, normalization, or log scaling) or whether to quantize the feature and treat it like a categorical variable. 
While we have not explored this idea fully, the best strategy is likely to use all the different types of encoding in parallel, turning each scalar feature into three re-scaled features and one categorical feature.  Unlike learning embeddings for high-cardinality categorical features, adding potentially-redundant encodings for scalar variables should not lead to overfitting, but can make the difference between a feature being useful or not.

For text variables, we simply encodes the number of words and character in the text.

\section{Appendix: Benchmark Dataset Information and Experiment Results}\label{sup:experiment_results}

\begin{table*}[t]
\caption{Benchmark datasets.  All datasets are binary classification tasks. Positive Class\% is the fraction of data points that belongs to the positive class.}
\label{tab:dataset_info}
\centering
\scalebox{0.9}{
\begin{tabular}{lllll}
\toprule
    Dataset Name &  N Datapoints &  N Features &   Positive Class\% \\
\midrule
     1995\_income &         32561 &          14 &                         24.1 \\
          adult &         34190 &          25 &                         85.4 \\
          albert &        425240 &          79 &                         50.0 \\
  bank\_marketing &         45211 &          16 &                         11.7 \\
      blastchar &          7043 &          20 &                         26.5 \\
      dota2games &         92650 &         117 &                         52.7 \\
          fabert &          8237 &         801 &                         11.3 \\
      hcdr\_main &        307511 &         120 &                          8.1 \\
          htru2 &         17898 &           8 &                          9.2 \\
    insurance\_co &          5822 &          85 &                          6.0 \\
          jannis &         83733 &          55 &                          2.0 \\
         jasmine &          2984 &         145 &                         50.0 \\
 online\_shoppers &         12330 &          17 &                         15.5 \\
      philippine &          5832 &         309 &                         50.0 \\
        qsar\_bio &          1055 &          41 &                         33.7 \\
    seismicbumps &          2583 &          18 &                          6.6 \\
        shrutime &         10000 &          11 &                         20.4 \\
        spambase &          4601 &          57 &                         39.4 \\
         sylvine &          5124 &          20 &                         50.0 \\
         volkert &         58310 &         181 &                         12.7 \\
\bottomrule
\end{tabular}}
\end{table*}

\begin{table*}[t]
\caption{Benchmark Dataset Links.}
\label{sup:tab:dataset_urls}
\centering
\scalebox{0.81}{
\begin{tabular}{ll}
\toprule
    Dataset Name &                                                                                         URL \\
\midrule
     1995\_income &                           \url{https://www.kaggle.com/lodetomasi1995/income-classification} \\
           adult &                                                       \url{http://automl.chalearn.org/data} \\
          albert &                                                       \url{http://automl.chalearn.org/data} \\
  bank\_marketing &                                \url{https://archive.ics.uci.edu/ml/datasets/bank+marketing} \\
       blastchar &                                 \url{https://www.kaggle.com/blastchar/telco-customer-churn} \\
      dota2games &                           \url{https://archive.ics.uci.edu/ml/datasets/Dota2+Games+Results} \\
          fabert &                                                       \url{http://automl.chalearn.org/data} \\
       hcdr\_main &                                     \url{https://www.kaggle.com/c/home-credit-default-risk} \\
           htru2 &                                         \url{https://archive.ics.uci.edu/ml/datasets/HTRU2} \\
    insurance\_co &   \url{https://archive.ics.uci.edu/ml/datasets/Insurance+Company+Benchmark+\%28COIL+2000\%29} \\
          jannis &                                                       \url{http://automl.chalearn.org/data} \\
         jasmine &                                                       \url{http://automl.chalearn.org/data} \\
 online\_shoppers &  \url{https://archive.ics.uci.edu/ml/datasets/Online+Shoppers+Purchasing+Intention+Dataset} \\
      philippine &                                                       \url{http://automl.chalearn.org/data} \\
        qsar\_bio &                           \url{https://archive.ics.uci.edu/ml/datasets/QSAR+biodegradation} \\
    seismicbumps &                                 \url{https://archive.ics.uci.edu/ml/datasets/seismic-bumps} \\
        shrutime &                                \url{https://www.kaggle.com/shrutimechlearn/churn-modelling} \\
        spambase &                                      \url{https://archive.ics.uci.edu/ml/datasets/Spambase} \\
         sylvine &                                                       \url{http://automl.chalearn.org/data} \\
         volkert &                                                       \url{http://automl.chalearn.org/data} \\
\bottomrule
\end{tabular}
}
\end{table*}

\begin{table*}[t]
\caption{AUC score for semi-supervised learning models on all datasets with {\bf50} fine-tune data points. Values are the mean over 5 cross-validation splits, plus or minus the standard deviation. Larger values means better result. }
\centering
\label{sup:tab:semisup-result-auroc-50-1}
\tabcolsep=0.11cm
\scalebox{0.65}{
\begin{tabular}{llllllll}
\toprule
Dataset &  N Datapoints &  N Features &  Positive Class\% &                       Best Model & TabTransformer-RTD & TabTransformer-MLM & MLP (ER) \\

\midrule
albert          &        425240 &          79 &                         50.0 &             TabTransformer-MLM &    0.644 $\pm$ 0.015 &    0.647 $\pm$ 0.019 &   0.612 $\pm$ 0.017 \\
hcdr\_main       &        307511 &         120 &                          8.1 &                         MLP (DAE) &    0.592 $\pm$ 0.047 &    0.596 $\pm$ 0.047 &   0.602 $\pm$ 0.033 \\
dota2games      &         92650 &         117 &                         52.7 &             TabTransformer-MLM &    0.526 $\pm$ 0.009 &    0.538 $\pm$ 0.011 &   0.519 $\pm$ 0.007 \\
jannis          &         83733 &          55 &                          2.0 &             TabTransformer-RTD &    0.684 $\pm$ 0.055 &    0.665 $\pm$ 0.056 &   0.621 $\pm$ 0.022 \\
volkert         &         58310 &         181 &                          1.0 &             TabTransformer-RTD &    0.693 $\pm$ 0.046 &    0.689 $\pm$ 0.042 &   0.657 $\pm$ 0.028 \\
bank\_marketing  &         45211 &          16 &                         11.7 &                              MLP (PL) &    0.771 $\pm$ 0.046 &    0.735 $\pm$ 0.040 &   0.792 $\pm$ 0.039 \\
adult           &         34190 &          25 &                         85.4 &                         MLP (DAE) &    0.580 $\pm$ 0.012 &    0.613 $\pm$ 0.014 &   0.609 $\pm$ 0.005 \\
1995\_income     &         32561 &          14 &                         24.1 &             TabTransformer-MLM &    0.840 $\pm$ 0.029 &    0.862 $\pm$ 0.018 &   0.839 $\pm$ 0.034 \\
htru2           &         17898 &           8 &                          9.2 &                         MLP (DAE) &    0.956 $\pm$ 0.007 &    0.958 $\pm$ 0.009 &   0.969 $\pm$ 0.012 \\
online\_shoppers &         12330 &          17 &                         15.5 &                              MLP (DAE) &    0.790 $\pm$ 0.013 &    0.780 $\pm$ 0.024 &   0.855 $\pm$ 0.019 \\
shrutime        &         10000 &          11 &                         20.4 &             TabTransformer-RTD &    0.752 $\pm$ 0.019 &    0.741 $\pm$ 0.019 &   0.725 $\pm$ 0.032 \\
fabert          &          8237 &         801 &                         11.3 &                 MLP (PL) &    0.535 $\pm$ 0.027 &    0.525 $\pm$ 0.019 &   0.572 $\pm$ 0.019 \\
blastchar       &          7043 &          20 &                         26.5 &             TabTransformer-MLM &    0.806 $\pm$ 0.018 &    0.822 $\pm$ 0.009 &   0.803 $\pm$ 0.021 \\
philippine      &          5832 &         309 &                         50.0 &             TabTransformer-RTD &    0.739 $\pm$ 0.027 &    0.729 $\pm$ 0.035 &   0.722 $\pm$ 0.031 \\
insurance\_co    &          5822 &          85 &                          6.0 &                 MLP (PL) &    0.601 $\pm$ 0.056 &    0.573 $\pm$ 0.077 &   0.575 $\pm$ 0.063 \\
sylvine         &          5124 &          20 &                         50.0 &                             MLP (PL) &    0.872 $\pm$ 0.031 &    0.898 $\pm$ 0.030 &   0.930 $\pm$ 0.015 \\
spambase        &          4601 &          57 &                         39.4 &              MLP (ER) &    0.949 $\pm$ 0.005 &    0.945 $\pm$ 0.011 &   0.957 $\pm$ 0.008 \\
jasmine         &          2984 &         145 &                         50.0 &             TabTransformer-MLM &    0.821 $\pm$ 0.019 &    0.837 $\pm$ 0.019 &   0.830 $\pm$ 0.022 \\
seismicbumps    &          2583 &          18 &                          6.6 &  TabTransformer (ER) &    0.740 $\pm$ 0.088 &    0.738 $\pm$ 0.068 &   0.712 $\pm$ 0.074 \\
qsar\_bio        &          1055 &          41 &                         33.7 &                         MLP (DAE) &    0.875 $\pm$ 0.028 &    0.869 $\pm$ 0.036 &   0.880 $\pm$ 0.022 \\
\bottomrule
\end{tabular}
}
\end{table*}

\begin{table*}[t]
\caption{(Continued) AUC score for semi-supervised learning models on all datasets with {\bf50} fine-tune data points. Values are the mean over 5 cross-validation splits, plus or minus the standard deviation. Larger values means better result. }
\centering
\label{sup:tab:semisup-result-auroc-50-2}
\tabcolsep=0.11cm
\scalebox{0.7}{
\begin{tabular}{lccccc}
\toprule
Dataset &   MLP (PL) & TabTransformer (ER) & TabTransformer (PL) &           MLP (DAE) &  GBDT (PL) \\
\midrule
albert          &  0.607 $\pm$ 0.013 &               0.580 $\pm$ 0.017 &            0.587 $\pm$ 0.012 &  0.612 $\pm$ 0.014 &  0.547 $\pm$ 0.032 \\
hcdr\_main       &  0.599 $\pm$ 0.038 &               0.581 $\pm$ 0.023 &            0.570 $\pm$ 0.031 &  0.620 $\pm$ 0.028 &  0.531 $\pm$ 0.024 \\
dota2games      &  0.520 $\pm$ 0.006 &               0.516 $\pm$ 0.009 &            0.519 $\pm$ 0.008 &  0.516 $\pm$ 0.004 &  0.505 $\pm$ 0.008 \\
jannis          &  0.623 $\pm$ 0.035 &               0.582 $\pm$ 0.035 &            0.604 $\pm$ 0.013 &  0.626 $\pm$ 0.023 &  0.519 $\pm$ 0.047 \\
volkert         &  0.653 $\pm$ 0.035 &               0.635 $\pm$ 0.024 &            0.639 $\pm$ 0.040 &  0.629 $\pm$ 0.019 &  0.525 $\pm$ 0.018 \\
bank\_marketing  &  0.805 $\pm$ 0.036 &               0.744 $\pm$ 0.063 &            0.767 $\pm$ 0.058 &  0.786 $\pm$ 0.055 &  0.688 $\pm$ 0.057 \\
adult           &  0.605 $\pm$ 0.021 &               0.568 $\pm$ 0.012 &            0.582 $\pm$ 0.024 &  0.616 $\pm$ 0.010 &  0.519 $\pm$ 0.024 \\
1995\_income     &  0.819 $\pm$ 0.042 &               0.813 $\pm$ 0.045 &            0.822 $\pm$ 0.048 &  0.811 $\pm$ 0.042 &  0.685 $\pm$ 0.084 \\
htru2           &  0.970 $\pm$ 0.012 &               0.955 $\pm$ 0.007 &            0.951 $\pm$ 0.009 &  0.973 $\pm$ 0.003 &  0.919 $\pm$ 0.021 \\
online\_shoppers &  0.848 $\pm$ 0.021 &               0.816 $\pm$ 0.036 &            0.818 $\pm$ 0.028 &  0.858 $\pm$ 0.019 &  0.818 $\pm$ 0.032 \\
shrutime        &  0.715 $\pm$ 0.044 &               0.748 $\pm$ 0.035 &            0.739 $\pm$ 0.034 &  0.683 $\pm$ 0.055 &  0.651 $\pm$ 0.093 \\
fabert          &  0.577 $\pm$ 0.027 &               0.504 $\pm$ 0.020 &            0.516 $\pm$ 0.020 &  0.552 $\pm$ 0.013 &  0.534 $\pm$ 0.016 \\
blastchar       &  0.799 $\pm$ 0.025 &               0.799 $\pm$ 0.013 &            0.792 $\pm$ 0.025 &  0.817 $\pm$ 0.016 &  0.729 $\pm$ 0.053 \\
philippine      &  0.725 $\pm$ 0.022 &               0.689 $\pm$ 0.046 &            0.703 $\pm$ 0.050 &  0.717 $\pm$ 0.022 &  0.628 $\pm$ 0.085 \\
insurance\_co    &  0.601 $\pm$ 0.057 &               0.575 $\pm$ 0.066 &            0.592 $\pm$ 0.080 &  0.522 $\pm$ 0.052 &  0.560 $\pm$ 0.081 \\
sylvine         &  0.939 $\pm$ 0.013 &               0.891 $\pm$ 0.022 &            0.904 $\pm$ 0.027 &  0.925 $\pm$ 0.010 &  0.914 $\pm$ 0.021 \\
spambase        &  0.951 $\pm$ 0.010 &               0.947 $\pm$ 0.006 &            0.948 $\pm$ 0.006 &  0.949 $\pm$ 0.012 &  0.899 $\pm$ 0.039 \\
jasmine         &  0.819 $\pm$ 0.021 &               0.825 $\pm$ 0.024 &            0.819 $\pm$ 0.018 &  0.812 $\pm$ 0.029 &  0.755 $\pm$ 0.016 \\
seismicbumps    &  0.678 $\pm$ 0.106 &               0.745 $\pm$ 0.080 &            0.713 $\pm$ 0.090 &  0.724 $\pm$ 0.049 &  0.601 $\pm$ 0.071 \\
qsar\_bio        &  0.875 $\pm$ 0.015 &               0.851 $\pm$ 0.041 &            0.835 $\pm$ 0.053 &  0.888 $\pm$ 0.022 &  0.804 $\pm$ 0.057 \\
\bottomrule
\end{tabular}
}
\end{table*}

\begin{table*}[t]
\caption{AUC score for semi-supervised learning models on all datasets with {\bf200} fine-tune data points. Values are the mean over 5 cross-validation splits, plus or minus the standard deviation. Larger values means better result.  }
\centering
\label{sup:tab:semisup-result-auroc-200-1}
\tabcolsep=0.11cm
\scalebox{0.65}{
\begin{tabular}{llllllll}
\toprule
Dataset &  N Datapoints &  N Features &  Positive Class\% &                       Best Model & TabTransformer-RTD & TabTransformer-MLM & MLP (ER) \\
\midrule
albert          &        425240 &          79 &                         50.0 &             TabTransformer-MLM &    0.699 $\pm$ 0.011 &    0.701 $\pm$ 0.014 &   0.642 $\pm$ 0.020 \\
hcdr\_main       &        307511 &         120 &                          8.1 &             TabTransformer-MLM &    0.655 $\pm$ 0.040 &    0.668 $\pm$ 0.028 &   0.639 $\pm$ 0.027 \\
dota2games      &         92650 &         117 &                         52.7 &             TabTransformer-MLM &    0.536 $\pm$ 0.012 &    0.549 $\pm$ 0.008 &   0.527 $\pm$ 0.012 \\
jannis          &         83733 &          55 &                          2.0 &             TabTransformer-RTD &    0.713 $\pm$ 0.037 &    0.692 $\pm$ 0.024 &   0.665 $\pm$ 0.024 \\
volkert         &         58310 &         181 &                         12.7 &             TabTransformer-RTD &    0.753 $\pm$ 0.022 &    0.742 $\pm$ 0.023 &   0.696 $\pm$ 0.033 \\
bank\_marketing  &         45211 &          16 &                         11.7 &                              MLP (PL) &    0.854 $\pm$ 0.020 &    0.838 $\pm$ 0.010 &   0.860 $\pm$ 0.008 \\
adult           &         34190 &          25 &                         85.4 &                              MLP (ER) &    0.596 $\pm$ 0.023 &    0.614 $\pm$ 0.012 &   0.623 $\pm$ 0.017 \\
1995\_income     &         32561 &          14 &                         24.1 &             TabTransformer-MLM &    0.866 $\pm$ 0.014 &    0.875 $\pm$ 0.011 &   0.868 $\pm$ 0.007 \\
htru2           &         17898 &           8 &                          9.2 &                              MLP (DAE) &    0.961 $\pm$ 0.008 &    0.963 $\pm$ 0.009 &   0.974 $\pm$ 0.007 \\
online\_shoppers &         12330 &          17 &                         15.5 &                             MLP (ER) &    0.834 $\pm$ 0.015 &    0.838 $\pm$ 0.024 &   0.876 $\pm$ 0.019 \\
shrutime        &         10000 &          11 &                         20.4 &             TabTransformer-RTD &    0.805 $\pm$ 0.017 &    0.783 $\pm$ 0.024 &   0.773 $\pm$ 0.013 \\
fabert          &          8237 &         801 &                         11.3 &              MLP (ER) &    0.556 $\pm$ 0.023 &    0.561 $\pm$ 0.028 &   0.600 $\pm$ 0.046 \\
blastchar       &          7043 &          20 &                         26.5 &             TabTransformer-MLM &    0.831 $\pm$ 0.010 &    0.841 $\pm$ 0.014 &   0.829 $\pm$ 0.010 \\
philippine      &          5832 &         309 &                         50.0 &             TabTransformer-RTD &    0.757 $\pm$ 0.017 &    0.754 $\pm$ 0.016 &   0.732 $\pm$ 0.024 \\
insurance\_co    &          5822 &          85 &                          6.0 &  TabTransformer (ER) &    0.667 $\pm$ 0.062 &    0.640 $\pm$ 0.043 &   0.601 $\pm$ 0.059 \\
sylvine         &          5124 &          20 &                         50.0 &                 MLP (PL) &    0.939 $\pm$ 0.008 &    0.948 $\pm$ 0.006 &   0.957 $\pm$ 0.008 \\
spambase        &          4601 &          57 &                         39.4 &              MLP (ER) &    0.957 $\pm$ 0.006 &    0.955 $\pm$ 0.010 &   0.968 $\pm$ 0.009 \\
jasmine         &          2984 &         145 &                         50.0 &             TabTransformer-RTD &    0.843 $\pm$ 0.016 &    0.843 $\pm$ 0.028 &   0.831 $\pm$ 0.019 \\
seismicbumps    &          2583 &          18 &                          6.6 &             TabTransformer-RTD &    0.738 $\pm$ 0.063 &    0.708 $\pm$ 0.083 &   0.694 $\pm$ 0.088 \\
qsar\_bio        &          1055 &          41 &                         33.7 &             TabTransformer-RTD &    0.896 $\pm$ 0.018 &    0.889 $\pm$ 0.030 &   0.895 $\pm$ 0.026 \\
\bottomrule
\end{tabular}
}
\end{table*}

\begin{table*}[t]
\caption{(Continued) AUC score for semi-supervised learning models on all datasets with {\bf200} fine-tune data points. Values are the mean over 5 cross-validation splits, plus or minus the standard deviation. Larger values means better result.  }
\centering
\label{sup:tab:semisup-result-auroc-200-2}
\tabcolsep=0.11cm
\scalebox{0.7}{
\begin{tabular}{lccccc}
\toprule
Dataset &   MLP (PL) & TabTransformer (ER) & TabTransformer (PL) &           MLP (DAE) &  GBDT (PL) \\
\midrule
albert          &  0.638 $\pm$ 0.024 &               0.630 $\pm$ 0.025 &            0.630 $\pm$ 0.021 &  0.646 $\pm$ 0.023 &  0.628 $\pm$ 0.015 \\
hcdr\_main       &  0.631 $\pm$ 0.019 &               0.611 $\pm$ 0.030 &            0.605 $\pm$ 0.021 &  0.636 $\pm$ 0.027 &  0.579 $\pm$ 0.039 \\
dota2games      &  0.527 $\pm$ 0.014 &               0.528 $\pm$ 0.017 &            0.525 $\pm$ 0.011 &  0.528 $\pm$ 0.012 &  0.506 $\pm$ 0.008 \\
jannis          &  0.667 $\pm$ 0.036 &               0.619 $\pm$ 0.024 &            0.637 $\pm$ 0.026 &  0.659 $\pm$ 0.020 &  0.525 $\pm$ 0.030 \\
volkert         &  0.693 $\pm$ 0.028 &               0.694 $\pm$ 0.002 &            0.689 $\pm$ 0.015 &  0.672 $\pm$ 0.015 &  0.612 $\pm$ 0.042 \\
bank\_marketing  &  0.866 $\pm$ 0.008 &               0.853 $\pm$ 0.016 &            0.858 $\pm$ 0.009 &  0.863 $\pm$ 0.009 &  0.802 $\pm$ 0.012 \\
adult           &  0.616 $\pm$ 0.014 &               0.582 $\pm$ 0.026 &            0.584 $\pm$ 0.017 &  0.611 $\pm$ 0.027 &  0.572 $\pm$ 0.040 \\
1995\_income     &  0.869 $\pm$ 0.009 &               0.848 $\pm$ 0.024 &            0.852 $\pm$ 0.015 &  0.865 $\pm$ 0.011 &  0.822 $\pm$ 0.020 \\
htru2           &  0.974 $\pm$ 0.007 &               0.955 $\pm$ 0.007 &            0.954 $\pm$ 0.007 &  0.974 $\pm$ 0.010 &  0.946 $\pm$ 0.022 \\
online\_shoppers &  0.873 $\pm$ 0.030 &               0.857 $\pm$ 0.014 &            0.853 $\pm$ 0.017 &  0.873 $\pm$ 0.021 &  0.846 $\pm$ 0.019 \\
shrutime        &  0.774 $\pm$ 0.018 &               0.803 $\pm$ 0.022 &            0.803 $\pm$ 0.024 &  0.763 $\pm$ 0.018 &  0.750 $\pm$ 0.050 \\
fabert          &  0.595 $\pm$ 0.048 &               0.530 $\pm$ 0.027 &            0.522 $\pm$ 0.024 &  0.580 $\pm$ 0.020 &  0.573 $\pm$ 0.026 \\
blastchar       &  0.829 $\pm$ 0.011 &               0.823 $\pm$ 0.011 &            0.823 $\pm$ 0.011 &  0.832 $\pm$ 0.013 &  0.783 $\pm$ 0.017 \\
philippine      &  0.733 $\pm$ 0.018 &               0.736 $\pm$ 0.018 &            0.739 $\pm$ 0.024 &  0.720 $\pm$ 0.020 &  0.729 $\pm$ 0.024 \\
insurance\_co    &  0.616 $\pm$ 0.045 &               0.715 $\pm$ 0.038 &            0.680 $\pm$ 0.034 &  0.612 $\pm$ 0.024 &  0.630 $\pm$ 0.087 \\
sylvine         &  0.961 $\pm$ 0.004 &               0.951 $\pm$ 0.009 &            0.950 $\pm$ 0.010 &  0.955 $\pm$ 0.009 &  0.957 $\pm$ 0.005 \\
spambase        &  0.965 $\pm$ 0.008 &               0.962 $\pm$ 0.006 &            0.960 $\pm$ 0.008 &  0.964 $\pm$ 0.009 &  0.957 $\pm$ 0.013 \\
jasmine         &  0.839 $\pm$ 0.013 &               0.824 $\pm$ 0.024 &            0.841 $\pm$ 0.016 &  0.842 $\pm$ 0.014 &  0.826 $\pm$ 0.013 \\
seismicbumps    &  0.684 $\pm$ 0.071 &               0.723 $\pm$ 0.080 &            0.727 $\pm$ 0.081 &  0.673 $\pm$ 0.070 &  0.603 $\pm$ 0.023 \\
qsar\_bio        &  0.892 $\pm$ 0.033 &               0.871 $\pm$ 0.036 &            0.876 $\pm$ 0.032 &  0.891 $\pm$ 0.018 &  0.855 $\pm$ 0.035 \\
\bottomrule
\end{tabular}
}
\end{table*}

\begin{table*}[t]
\caption{AUC score for semi-supervised learning models on all datasets with {\bf500} fine-tune data points. Values are the mean over 5 cross-validation splits, plus or minus the standard deviation. Larger values means better result.  }
\label{tab:semisup-result-auroc-500-1}
\tabcolsep=0.11cm
\centering
\scalebox{0.65}{
\begin{tabular}{llllllll}
\toprule
Dataset &  N Datapoints &  N Features & Positive Class\% &                       Best Model & TabTransformer-RTD & TabTransformer-MLM & MLP (ER) \\
\midrule
albert          &        425240 &          79 &                         50.0 &             TabTransformer-RTD &    0.711 $\pm$ 0.004 &    0.707 $\pm$ 0.006 &   0.666 $\pm$ 0.008 \\
hcdr\_main       &        307511 &         120 &                          8.1 &             TabTransformer-MLM &    0.690 $\pm$ 0.038 &    0.698 $\pm$ 0.033 &   0.653 $\pm$ 0.019 \\
dota2games      &         92650 &         117 &                         52.7 &             TabTransformer-MLM &    0.548 $\pm$ 0.008 &    0.557 $\pm$ 0.003 &   0.543 $\pm$ 0.008 \\
jannis          &         83733 &          55 &                          2.0 &             TabTransformer-RTD &    0.747 $\pm$ 0.015 &    0.720 $\pm$ 0.018 &   0.707 $\pm$ 0.036 \\
volkert         &         58310 &         181 &                         12.7 &             TabTransformer-RTD &    0.771 $\pm$ 0.016 &    0.760 $\pm$ 0.015 &   0.723 $\pm$ 0.016 \\
bank\_marketing  &         45211 &          16 &                         11.7 &             TabTransformer-RTD &    0.879 $\pm$ 0.012 &    0.866 $\pm$ 0.016 &   0.869 $\pm$ 0.012 \\
adult           &         34190 &          25 &                         85.4 &                              MLP (PL) &    0.625 $\pm$ 0.011 &    0.647 $\pm$ 0.008 &   0.644 $\pm$ 0.015 \\
1995\_income     &         32561 &          14 &                         24.1 &                         MLP (DAE) &    0.874 $\pm$ 0.008 &    0.880 $\pm$ 0.007 &   0.878 $\pm$ 0.002 \\
htru2           &         17898 &           8 &                          9.2 &                         MLP (DAE) &    0.964 $\pm$ 0.009 &    0.966 $\pm$ 0.009 &   0.973 $\pm$ 0.010 \\
online\_shoppers &         12330 &          17 &                         15.5 &                             MLP (ER) &    0.859 $\pm$ 0.009 &    0.861 $\pm$ 0.014 &   0.888 $\pm$ 0.012 \\
shrutime        &         10000 &          11 &                         20.4 &             TabTransformer-RTD &    0.831 $\pm$ 0.017 &    0.815 $\pm$ 0.004 &   0.793 $\pm$ 0.017 \\
fabert          &          8237 &         801 &                         11.3 &                              MLP (ER) &    0.618 $\pm$ 0.014 &    0.609 $\pm$ 0.019 &   0.621 $\pm$ 0.032 \\
blastchar       &          7043 &          20 &                         26.5 &             TabTransformer-RTD &    0.840 $\pm$ 0.013 &    0.839 $\pm$ 0.015 &   0.829 $\pm$ 0.013 \\
philippine      &          5832 &         309 &                         50.0 &             TabTransformer-MLM &    0.769 $\pm$ 0.028 &    0.772 $\pm$ 0.017 &   0.734 $\pm$ 0.024 \\
insurance\_co    &          5822 &          85 &                          6.0 &  TabTransformer (ER) &    0.688 $\pm$ 0.039 &    0.642 $\pm$ 0.029 &   0.659 $\pm$ 0.023 \\
sylvine         &          5124 &          20 &                         50.0 &                             MLP (PL) &    0.955 $\pm$ 0.007 &    0.959 $\pm$ 0.006 &   0.967 $\pm$ 0.003 \\
spambase        &          4601 &          57 &                         39.4 &                             MLP (ER) &    0.966 $\pm$ 0.007 &    0.968 $\pm$ 0.008 &   0.975 $\pm$ 0.004 \\
jasmine         &          2984 &         145 &                         50.0 &                             TabTransformer-RTD &    0.847 $\pm$ 0.016 &    0.844 $\pm$ 0.011 &   0.837 $\pm$ 0.019 \\
seismicbumps    &          2583 &          18 &                          6.6 &             TabTransformer-RTD &    0.758 $\pm$ 0.081 &    0.729 $\pm$ 0.069 &   0.682 $\pm$ 0.123 \\
qsar\_bio        &          1055 &          41 &                         33.7 &                         MLP (DAE) &    0.909 $\pm$ 0.024 &    0.889 $\pm$ 0.038 &   0.918 $\pm$ 0.023 \\
\bottomrule
\end{tabular}
}
\end{table*}

\begin{table*}[t]
\caption{(Continued) AUC score for semi-supervised learning models on all datasets with {\bf500} fine-tune data points. Values are the mean over 5 cross-validation splits, plus or minus the standard deviation. Larger values means better result.  }
\label{tab:semisup-result-auroc-500-2}
\tabcolsep=0.11cm
\centering
\scalebox{0.7}{
\begin{tabular}{lccccc}
\toprule
Dataset &   MLP (PL) & TabTransformer (ER) & TabTransformer (PL) &           MLP (DAE) &  GBDT (PL) \\
\midrule
albert          &  0.662 $\pm$ 0.007 &               0.664 $\pm$ 0.011 &            0.643 $\pm$ 0.029 &  0.666 $\pm$ 0.006 &  0.653 $\pm$ 0.011 \\
hcdr\_main       &  0.645 $\pm$ 0.022 &               0.623 $\pm$ 0.036 &            0.636 $\pm$ 0.031 &  0.657 $\pm$ 0.033 &  0.607 $\pm$ 0.035 \\
dota2games      &  0.544 $\pm$ 0.010 &               0.538 $\pm$ 0.009 &            0.541 $\pm$ 0.010 &  0.542 $\pm$ 0.012 &  0.505 $\pm$ 0.005 \\
jannis          &  0.698 $\pm$ 0.033 &               0.662 $\pm$ 0.007 &            0.660 $\pm$ 0.024 &  0.693 $\pm$ 0.024 &  0.521 $\pm$ 0.045 \\
volkert         &  0.722 $\pm$ 0.012 &               0.712 $\pm$ 0.016 &            0.705 $\pm$ 0.021 &  0.712 $\pm$ 0.016 &  0.705 $\pm$ 0.016 \\
bank\_marketing  &  0.876 $\pm$ 0.017 &               0.863 $\pm$ 0.008 &            0.868 $\pm$ 0.016 &  0.874 $\pm$ 0.012 &  0.838 $\pm$ 0.019 \\
adult           &  0.651 $\pm$ 0.012 &               0.618 $\pm$ 0.023 &            0.618 $\pm$ 0.021 &  0.654 $\pm$ 0.016 &  0.647 $\pm$ 0.030 \\
1995\_income     &  0.880 $\pm$ 0.003 &               0.868 $\pm$ 0.008 &            0.869 $\pm$ 0.007 &  0.882 $\pm$ 0.001 &  0.839 $\pm$ 0.013 \\
htru2           &  0.974 $\pm$ 0.007 &               0.960 $\pm$ 0.010 &            0.960 $\pm$ 0.008 &  0.976 $\pm$ 0.006 &  0.949 $\pm$ 0.007 \\
online\_shoppers &  0.885 $\pm$ 0.021 &               0.861 $\pm$ 0.011 &            0.860 $\pm$ 0.013 &  0.885 $\pm$ 0.019 &  0.865 $\pm$ 0.011 \\
shrutime        &  0.800 $\pm$ 0.015 &               0.825 $\pm$ 0.013 &            0.822 $\pm$ 0.016 &  0.804 $\pm$ 0.015 &  0.788 $\pm$ 0.019 \\
fabert          &  0.596 $\pm$ 0.046 &               0.573 $\pm$ 0.048 &            0.578 $\pm$ 0.033 &  0.617 $\pm$ 0.042 &  0.585 $\pm$ 0.025 \\
blastchar       &  0.833 $\pm$ 0.013 &               0.834 $\pm$ 0.013 &            0.832 $\pm$ 0.011 &  0.833 $\pm$ 0.012 &  0.795 $\pm$ 0.021 \\
philippine      &  0.740 $\pm$ 0.023 &               0.746 $\pm$ 0.020 &            0.735 $\pm$ 0.015 &  0.739 $\pm$ 0.017 &  0.749 $\pm$ 0.026 \\
insurance\_co    &  0.646 $\pm$ 0.048 &               0.710 $\pm$ 0.040 &            0.666 $\pm$ 0.060 &  0.612 $\pm$ 0.013 &  0.672 $\pm$ 0.037 \\
sylvine         &  0.968 $\pm$ 0.003 &               0.958 $\pm$ 0.005 &            0.958 $\pm$ 0.003 &  0.967 $\pm$ 0.003 &  0.967 $\pm$ 0.006 \\
spambase        &  0.973 $\pm$ 0.005 &               0.968 $\pm$ 0.007 &            0.967 $\pm$ 0.006 &  0.972 $\pm$ 0.006 &  0.972 $\pm$ 0.005 \\
jasmine         &  0.833 $\pm$ 0.009 &               0.833 $\pm$ 0.021 &            0.838 $\pm$ 0.018 &  0.842 $\pm$ 0.011 &  0.838 $\pm$ 0.022 \\
seismicbumps    &  0.677 $\pm$ 0.103 &               0.687 $\pm$ 0.100 &            0.735 $\pm$ 0.081 &  0.696 $\pm$ 0.112 &  0.666 $\pm$ 0.063 \\
qsar\_bio        &  0.914 $\pm$ 0.032 &               0.894 $\pm$ 0.036 &            0.895 $\pm$ 0.035 &  0.925 $\pm$ 0.034 &  0.908 $\pm$ 0.024 \\
\bottomrule
\end{tabular}
}
\end{table*}

\begin{table*}[t]
\caption{AUC score for supervised learning models on all datasets. Values are the mean over 5 cross-validation splits, plus or minus the standard deviation. Larger values means better result.  }
\label{tab:supervised-result-auroc1}
\centering
\scalebox{0.65}{
\begin{tabular}{lllllll}
\toprule
Dataset &  N Datapoints &  N Features &  Positive Class\% &           Best Model & Logistic Regression &               GBDT \\
ds\_name         &               &             &                              &                      &                     &                    \\
\midrule
albert          &        425240 &          79 &                         50.0 &                 GBDT &   0.726 $\pm$ 0.001 &  0.763 $\pm$ 0.001 \\
hcdr\_main       &        307511 &         120 &                          8.1 &                 GBDT &   0.747 $\pm$ 0.004 &  0.756 $\pm$ 0.004 \\
dota2games      &         92650 &         117 &                         52.7 &  Logistic Regression &   0.634 $\pm$ 0.003 &  0.621 $\pm$ 0.004 \\
bank\_marketing  &         45211 &          16 &                         11.7 &       TabTransformer &   0.911 $\pm$ 0.005 &  0.933 $\pm$ 0.003 \\
adult           &         34190 &          25 &                         85.4 &                 GBDT &   0.721 $\pm$ 0.010 &  0.756 $\pm$ 0.011 \\
1995\_income     &         32561 &          14 &                         24.1 &                 TabTransformer &   0.899 $\pm$ 0.002 &  0.906 $\pm$ 0.002 \\
online\_shoppers &         12330 &          17 &                         15.5 &                 GBDT &   0.908 $\pm$ 0.015 &  0.930 $\pm$ 0.008 \\
shrutime        &         10000 &          11 &                         20.4 &                 GBDT &   0.828 $\pm$ 0.013 &  0.859 $\pm$ 0.009 \\
blastchar       &          7043 &          20 &                         26.5 &                 GBDT &   0.844 $\pm$ 0.010 &  0.847 $\pm$ 0.016 \\
philippine      &          5832 &         309 &                         50.0 &                 TabTransformer &   0.725 $\pm$ 0.022 &  0.812 $\pm$ 0.013 \\
insurance\_co    &          5822 &          85 &                          6.0 &       TabTransformer &   0.736 $\pm$ 0.023 &  0.732 $\pm$ 0.022 \\
spambase        &          4601 &          57 &                         39.4 &                 GBDT &   0.947 $\pm$ 0.008 &  0.987 $\pm$ 0.005 \\
jasmine         &          2984 &         145 &                         50.0 &                 GBDT &   0.846 $\pm$ 0.017 &  0.862 $\pm$ 0.008 \\
seismicbumps    &          2583 &          18 &                          6.6 &                 GBDT &   0.749 $\pm$ 0.068 &  0.756 $\pm$ 0.084 \\
qsar\_bio        &          1055 &          41 &                         33.7 &       TabTransformer &   0.847 $\pm$ 0.037 &  0.913 $\pm$ 0.031 \\
\bottomrule
\end{tabular}}
\end{table*}

\begin{table*}[t]
\caption{(Continued) AUC score for supervised learning models on all  datasets. Values are the mean over 5 cross-validation splits, plus or minus the standard deviation. Larger values means better result.}
\label{tab:supervised-result-auroc2}
\centering
\scalebox{0.78}{
\begin{tabular}{llllll}
\toprule
{} &                MLP &         Sparse MLP &     TabTransformer &             TabNet &                VIB \\
ds\_name         &                    &                    &                    &                    &                    \\
\midrule
albert          &  0.740 $\pm$ 0.001 &  0.741 $\pm$ 0.001 &  0.757 $\pm$ 0.002 &  0.705 $\pm$ 0.005 &  0.737 $\pm$ 0.001 \\
hcdr\_main       &  0.743 $\pm$ 0.004 &  0.753 $\pm$ 0.004 &  0.751 $\pm$ 0.004 &  0.711 $\pm$ 0.006 &  0.745 $\pm$ 0.005 \\
dota2games      &  0.631 $\pm$ 0.002 &  0.633 $\pm$ 0.004 &  0.633 $\pm$ 0.002 &  0.529 $\pm$ 0.025 &  0.628 $\pm$ 0.003 \\
bank\_marketing  &  0.929 $\pm$ 0.003 &  0.926 $\pm$ 0.007 &  0.934 $\pm$ 0.004 &  0.885 $\pm$ 0.017 &  0.920 $\pm$ 0.005 \\
adult           &  0.725 $\pm$ 0.010 &  0.740 $\pm$ 0.007 &  0.737 $\pm$ 0.009 &  0.663 $\pm$ 0.016 &  0.733 $\pm$ 0.009 \\
1995\_income     &  0.905 $\pm$ 0.003 &  0.904 $\pm$ 0.004 &  0.906 $\pm$ 0.003 &  0.875 $\pm$ 0.006 &  0.904 $\pm$ 0.003 \\
online\_shoppers &  0.919 $\pm$ 0.010 &  0.922 $\pm$ 0.011 &  0.927 $\pm$ 0.010 &  0.888 $\pm$ 0.020 &  0.907 $\pm$ 0.012 \\
shrutime        &  0.846 $\pm$ 0.013 &  0.828 $\pm$ 0.007 &  0.856 $\pm$ 0.005 &  0.785 $\pm$ 0.024 &  0.833 $\pm$ 0.011 \\
blastchar       &  0.839 $\pm$ 0.010 &  0.842 $\pm$ 0.015 &  0.835 $\pm$ 0.014 &  0.816 $\pm$ 0.014 &  0.842 $\pm$ 0.012 \\
philippine      &  0.821 $\pm$ 0.020 &  0.764 $\pm$ 0.018 &  0.834 $\pm$ 0.018 &  0.721 $\pm$ 0.008 &  0.757 $\pm$ 0.018 \\
insurance\_co    &  0.697 $\pm$ 0.027 &  0.705 $\pm$ 0.054 &  0.744 $\pm$ 0.009 &  0.630 $\pm$ 0.061 &  0.647 $\pm$ 0.028 \\
spambase        &  0.984 $\pm$ 0.004 &  0.980 $\pm$ 0.009 &  0.985 $\pm$ 0.005 &  0.975 $\pm$ 0.008 &  0.983 $\pm$ 0.004 \\
jasmine         &  0.851 $\pm$ 0.015 &  0.856 $\pm$ 0.013 &  0.853 $\pm$ 0.015 &  0.816 $\pm$ 0.017 &  0.847 $\pm$ 0.017 \\
seismicbumps    &  0.735 $\pm$ 0.028 &  0.699 $\pm$ 0.074 &  0.751 $\pm$ 0.096 &  0.701 $\pm$ 0.051 &  0.681 $\pm$ 0.084 \\
qsar\_bio        &  0.910 $\pm$ 0.037 &  0.916 $\pm$ 0.036 &  0.918 $\pm$ 0.038 &  0.860 $\pm$ 0.038 &  0.914 $\pm$ 0.028 \\
\bottomrule
\end{tabular}}
\end{table*}

\end{document}